\documentclass{article} 
\usepackage{iclr2027_conference,times}

\usepackage{amsmath,amsfonts,bm}

\def\eqref#1{equation~\ref{#1}}

\def\1{\bm{1}}

\DeclareMathAlphabet{\mathsfit}{\encodingdefault}{\sfdefault}{m}{sl}
\SetMathAlphabet{\mathsfit}{bold}{\encodingdefault}{\sfdefault}{bx}{n}

\usepackage{hyperref}
\usepackage{url}
\usepackage{graphicx}   
\usepackage{booktabs}   
\usepackage{amsmath}

\title{Reasoning Concentrates Errors, \\ and Self-Consistency Never Notices}

\author{Asaad Althoubi \\
Department of Computer Science \\
Oklahoma State University \\
Stillwater, OK, USA \\
\texttt{aalthou@okstate.edu} \\
}

\iclrfinalcopy 
\begin{document}

\maketitle
\lhead{Preprint. Under review.}

\begin{abstract}
Self-consistency assumes that independent samples disagree when a model is unsure, so agreement is
evidence of correctness. Holding weights fixed and toggling only a reasoning mode, over five
benchmarks and 74,944 samples, we show that reasoning concentrates a model's \emph{errors}: the
probability that two independently drawn wrong answers coincide rises in all ten dataset--scale
comparisons ($p = 0.00098$), and in nine of nine after restricting both arms to the problems each
gets wrong. Where the answer space is unbounded, reasoning cuts the distinct answers produced to
0.43--0.65 of the non-reasoning count; where it is bounded, both arms hold an identical option set
and reasoning concentrates mass on it instead, which no positional prior can explain at fixed
weights. The aggregate cost is smaller than the mechanism predicts, because reasoning also shrinks
the set of problems where answer diversity can decide anything, in ten of ten cells and by
$2.7\times$; normalized for available headroom, both arms convert a quarter of it in domain.
Confidence weighting does not recover what is left. Across 280 method--dataset--model combinations
on eight models and five benchmarks, not one beats plain majority voting after correction; weighted
voting agrees with it on 98.5\% of problem--method pairs and is right 56.3\% of the time on the
rest; and a signal's direction can invert within fixed weights, with answer log-probability
predicting correctness when reasoning is off and \emph{error} when it is on. A learned
six-signal combination gains nothing out of domain. Confidence signals should be evaluated on
decisions, not on discrimination.
\end{abstract}

\section{Introduction}

Sampling a language model several times and returning the most common answer is among the most
widely used inference-time techniques in practice \citep{wang2023selfconsistency}. Its logic is
simple: a model that is uncertain will produce different answers on different samples, so agreement
is evidence of correctness, and a plurality vote discards the scatter.

Four recent studies report that this logic does not hold as reliably as its use implies.
The agreement correlates with correctness only weakly, and the best-agreeing model can be the worst
calibrated \citep{ding2026agree}. Majority voting reduces per-problem accuracy on the majority of
hard science problems, and cheap gating signals cannot identify which problems to skip
\citep{bahuguna2026backfires}. A model can produce the same wrong answer on every sample, and
consistency-based detectors fail on exactly those cases, at lower chance rates
\citep{tan2025consistent}. Separately, reinforcement learning with verifiable rewards collapses
the diversity of reasoning continuations a model will explore \citep{saha2026bodhi}.

\textbf{Three of these papers name the same missing piece in their own limitations sections.}
\citet{bahuguna2026backfires}, having found that agreement fails as a gating signal, writes that
its failure ``remains without a mechanism, and we state that as an open problem rather than
absorbing it.'' \citet{ding2026agree} writes: ``We do not study reasoning-trained models, whose
internal reasoning may already supply the calibration lift we measure for prompted CoT.''
\citet{tan2025consistent} write that the causes ``still require deeper investigation'' and that
``future work may construct controlled experiments to investigate the causes.''

The obstacle is identification. Reasoning models differ from non-reasoning models in training
data, post-training procedure, tokenizer, and scale simultaneously, so a comparison between them
cannot attribute anything to reasoning in particular. \citet{bahuguna2026backfires} attempted the
one control that would resolve this (toggling a hosted model's thinking mode) and found that the
provider accepted the flag and ignored it, producing identical output either way; he scopes the
result as remaining ``open for the open-weights route.''

\textbf{This paper takes this route.} Qwen3 serves thinking and non-thinking modes from identical
weights, selected by a chat-template flag \citep{qwen3}. We compare a model to itself at two
scales, with parameters, tokenizer, prompt, sampling parameters, and grader held fixed and only the
generation of a reasoning chain differing, across five benchmarks and 74,944 samples, alongside
six further models from three families.

\textbf{Reasoning concentrates errors.} We measure the probability that two independently drawn
\emph{incorrect} answers to the same problem coincide. Within fixed weights it rises in all ten
dataset--scale comparisons (exact sign test $p = 0.00098$), and in nine of nine after restricting
both conditions to the problems each gets wrong, a control that matters, because the two arms
differ sharply in accuracy and would otherwise average over different problem sets. Where the
answer space is unbounded, reasoning roughly halves the number of distinct answers produced (339
against 522 on MATH-500, from the same weights); where it is bounded, both arms hold an identical
set of four or ten keys and reasoning concentrates mass on it instead. The second case rules out
the leading alternative explanation, since a positional prior cannot explain why one arm
concentrates more than the other on the same options with the same weights.

\textbf{The aggregate cost is not what the mechanism predicts.} Measured as accuracy gained from
one sample to eight, self-consistency appears to help the non-reasoning arm three times as much.
That comparison is invalid: the reasoning arm sits near ceiling with a third of the headroom, and
normalized for what is available both arms convert almost exactly a quarter of it. Reasoning has
two opposing effects: it concentrates errors, which hurts voting where voting is decided by
answer diversity, and it shrinks that set of problems by a factor of 2.7, because it is wrong less
often. Both are real, they roughly cancel, and neither is visible in the headline number.

\textbf{What does not cancel is the failure of confidence-weighted aggregation.} Across 280 method--dataset--model combinations, seven confidence-weighted aggregation methods over eight models and five benchmarks, not one improves on plain majority voting after Holm--Bonferroni correction applied once across the family \citep{holm1979}. Five reach nominal
significance uncorrected, compared with 14.0 expected by chance. Weighted voting almost never changes the
answer, and when it does it is right 56.3\% of the time. Nor is signal \emph{direction} a property
of the signal: answer log-probability predicts correctness with reasoning disabled and predicts
\emph{error} with it enabled, at identical 8B weights. Finally, the output length, a strong
confidence signal here, at 0.965 AUROC on one benchmark, and previous work attributes to 
post-training reasoning \citep{devic2026trace} show no difference between the arms in open-ended
mathematics, and its strongest instance in the study belongs to a model without reasoning training
at all. The variable that reconciles the two results is the answer format, not training.

\textbf{What we do not claim.} The same-weights toggle is not our contribution: it has been used
before \citep{kumaran2026commitment} and we cite it as a method. We do not claim that error
concentration produces a measurable self-consistency penalty at this scale; our replacement measure
points the right way at $p = 0.10$ and is reported as supporting, not confirmatory. We also withdrew
a threefold accuracy-gain result that did not survive normalization for headroom.

\section{Relationship to prior work}
\label{sec:related}

\paragraph{A cluster of negative results.} \citet{tan2025consistent} name the phenomenon: drawing
fifteen samples per question, they define a \emph{self-consistent error} as a case where every
sample is semantically equivalent to the greedy response and all are wrong, and show semantic
entropy's detection AUROC falls to 0.4608 (below chance) against 0.8820 on inconsistent
errors, with such errors stable or increasing with model scale. \citet{ding2026agree} shows
agreement is not accuracy: across 265{,}000 samples at $K = 50$, self-consistency correlates with
majority-correctness at Spearman 0.20--0.59, and gpt-4.1 is wrong 48\% of the time on GPQA-Diamond
at agreement above 0.8 despite agreeing most; an option-shuffle control shows a meaningful share of
multiple-choice agreement is \emph{positional} rather than semantic (majority accuracy 0.50
$\rightarrow$ 0.29 under permutation). His measurements come from a graduate-course cohort's
independently submitted runs with model snapshots unlogged, which bounds how tightly the 48\%
figure can be read. \citet{bahuguna2026backfires} shows the downstream cost: majority voting
\emph{reduces} per-problem accuracy on 56.6\% and 65.7\% of GPQA-Diamond problems for two models,
three verifier-free gating signals all fail, and dissenting samples emit their answer at a median
margin of 20.52 nats as committed as agreeing ones. \citet{saha2026bodhi} locates a cause
upstream: RLVR collapses the candidate preference entropy at the branch points, semantically rather than
stylistically.

\paragraph{What none of them could measure.} \citet{saha2026bodhi} build their probe set from
only correct traces (all incorrect responses were removed), so they measure diversity
\emph{among correct paths}, at a truncated continuation rather than a final answer, contrasting two
flavors of reasoning post-training rather than reasoning and its absence. This paper measures the
answer distribution under repeated sampling with the weights held fixed: the locus
\citet{bahuguna2026backfires} identifies as inferred but unmeasured, in the population
\citet{saha2026bodhi} excludes, with control \citet{ding2026agree} and \citet{tan2025consistent}
could not run. We claim the measurement, not the method: \citet{kumaran2026commitment} applies a
same-weight toggle to a 235B model to study verbal confidence and abstention, but draws a single
response per trial and analyzes neither agreement nor length.

\paragraph{The length literature.} \citet{vashurin2025line} establish that uncertainty scores are
confounded by output length even in nominally length-normalized measures, and propose post-hoc
residual debiasing; their own limitations anticipate our regime (``in tasks such as multi-step
reasoning $\ldots$ a linear fit may be insufficient''). \citet{santilli2025revisiting} raise the
problem one level: when the \emph{correctness function} is length-biased too, AUROC rankings are
provably distorted: exact-match grading is what makes our length results readable.
\citet{devic2026trace} find trace length a useful zero-shot confidence signal (0.62--0.79 AUROC for
reasoning models against 0.56--0.60 for base models) and attribute the difference to reasoning
post-training; we reproduce their multiple-choice numbers and disagree with the generalization
(Section~\ref{sec:length}). Finally, chain-of-thought raises confidence more in wrong answers than in right
ones \citep{mcq2025selfconfident} and reasoning-trained models become more overconfident with deeper
reasoning \citep{reasoning2026knowdontknow}: reasoning makes errors more \emph{confident}, and we
show that it makes them more \emph{consistent}.

\section{Setup}
\label{sec:setup}

\textbf{Models.} Eight open-weight models across four families (Table~\ref{tab:models}): Qwen3-8B and Qwen3-32B, each in
thinking and non-thinking mode \citep{qwen3}; Olmo-3-Think-7B and Olmo-3-Think-32B \citep{olmo3};
and Gemma-3-12B-IT and Gemma-3-27B-IT \citep{gemma3}. \textbf{The two Qwen3 pairs are the
identification strategy}: the two arms of a pair share parameters, tokenizer, pre-training corpus,
and post-training recipe exactly, and differ only in whether a reasoning chain is generated. The
other four models establish that what the pairs show is not a Qwen3 artifact.

\textbf{Datasets.} Five benchmarks, 1{,}171 problems (Table~\ref{tab:datasets}), chosen to vary the domain and answer format
independently: MATH-500 (500 problems, open-ended, in-domain;
\citealp{hendrycks2021math,lightman2024verify}), AIME (90) and AMC (83), both open-ended and
out-of-domain, GPQA-Diamond (198, four-option; \citealp{rein2024gpqa}) and MMLU-Pro (300,
ten-option; \citealp{wang2024mmlupro}). Signals and selectors are fitted on MATH-500 and evaluated
on the other four, so every out-of-domain number is a genuine transfer, and crossing format with
domain is what lets Section~\ref{sec:mech} separate the two.

\textbf{Generation and grading.} Every model answers every problem $k = 8$ times independently,
at temperature 0.6, top-$p$ 0.95, and a 32{,}768-token budget, giving \textbf{74,944 records}, served by vLLM 0.27.1
\citep{kwon2023vllm}. Answers are graded by \textbf{exact match after normalization}: numeric
comparison on the mathematics sets, option-letter comparison on the multiple-choice ones. This is
load-bearing rather than conventional: \citet{santilli2025revisiting} show that a length-dependent
correctness function credits a length-correlated confidence measure for detecting the grader's own
mistakes, and an exact-match grader has no length sensitivity for such an error to correlate with.
The grader records which layer produced its verdict, so an unparseable response is distinguishable
from one that gives a value that the benchmark does not offer, and every accuracy is reported beside its
answer rate, following \citet{bahuguna2026backfires}.

\textbf{Confidence signals.} Twelve signals in six families, all computed from the model's own
output at no additional inference cost: probes (a last-token probe, zero-shot self-verification),
token entropy, log-probability (answer mean and min, trace mean), trace length, agreement
(self-consistency, entropy over the $k$ answers) and hedging (count, density) with an external
reward model scored alongside as a paid baseline. Every signal is oriented so that a higher value
means more likely correct, declared once in a specification table, and unit-tested, because getting
it wrong flips an AUROC from 0.72 to 0.28.

\textbf{Statistical protocol.} Confidence intervals are cluster bootstrap over \emph{problems},
not samples. Section~\ref{sec:down} applies Holm--Bonferroni once across the entire family of 280
combinations; correcting per model, as an earlier version did, under-corrects roughly sixfold in
exactly the analysis whose purpose is to survive a multiplicity objection. Claims about the toggle
are tested paired on the weight-pair with the exact sign test; claims about generalization across
the eight models use an unpaired exact Mann--Whitney, noting that eight models include two
weight-sharing pairs. Multiple-choice statistics are computed after excluding responses the grader
could not map to an offered option: Without this, one model shows 20 distinct keys on a four-option
benchmark and exactly 4 after, and because the dilution scales with each model's unmappable rate, it
does not cancel between models. Finally, sixteen numeric predictions were recorded in writing before the analyses that would test them,
including the threshold used in Section~\ref{sec:mech}; five did not hold, three of those forced a claim to be
rewritten or withdrawn, and Table~\ref{tab:prereg} gives the complete scorecard.

\section{Reasoning concentrates errors}
\label{sec:mech}

\textbf{What we measure.} Self-consistency depends not on accuracy but on \emph{answer diversity
among errors}: if a model's wrong answers are spread across many values, a plurality vote can
discard them; if they concentrate on one value, the vote ratifies it. For a problem $p$ with $n$
incorrect samples of which $n_a$ gives the particular wrong answer $a$, the \textbf{collision index}
is
\[
C(p) \;=\; \sum_a \frac{n_a\,(n_a - 1)}{n\,(n-1)},
\]
the probability that two independently drawn incorrect answers coincide, an unbiased
within-problem Herfindahl index. We average $C$ over problems with $n \geq 2$. Answer keys are
normalized and, on multiple-choice benchmarks, restricted to the offered options. A closely related concentration measure was proposed independently \citep{aaai.Li}; Appendix~\ref{app:matched} relates it to ours.

\begin{table}[!tb]
\caption{Collision among wrong answers, within fixed weights. Every one of the ten
dataset--scale comparisons favors reasoning (exact sign test $p = 2^{-10} = 0.00098$).
The last column is the number of distinct answer keys each arm produces.}
\label{tab:collision}
\begin{center}
\footnotesize
\begin{tabular}{llccrc}
\toprule
dataset & pair & reasoning on & reasoning off & $\Delta$ & distinct keys (on / off) \\
\midrule
MATH-500     & 8B  & 0.509 & 0.379 & $+0.130$ & 339 / 522 \\
MATH-500     & 32B & 0.487 & 0.297 & $+0.190$ & 334 / 542 \\
AIME         & 8B  & 0.263 & 0.141 & $+0.122$ & 143 / 292 \\
AIME         & 32B & 0.193 & 0.096 & $+0.097$ & 152 / 324 \\
AMC          & 8B  & 0.253 & 0.218 & $+0.035$ &  67 / 106 \\
AMC          & 32B & 0.594 & 0.163 & $+0.431$ &  52 / 122 \\
GPQA-Diamond & 8B  & 0.771 & 0.650 & $+0.121$ &   4 / 4   \\
GPQA-Diamond & 32B & 0.744 & 0.609 & $+0.135$ &   4 / 4   \\
MMLU-Pro     & 8B  & 0.748 & 0.601 & $+0.147$ &  10 / 10  \\
MMLU-Pro     & 32B & 0.771 & 0.590 & $+0.181$ &  10 / 10  \\
\bottomrule
\end{tabular}
\end{center}
\end{table}

\textbf{The primary contrast.} Table~\ref{tab:collision} reports the ten
within-weights comparisons; median $\Delta = +0.133$, range $+0.035$ to $+0.431$. Under the null that the thinking flag has no effect, ten concordant signs give an exact permutation probability of
$2^{-10} = 0.00098$. We lead with the conservative reading rather than the $p$-value: \textbf{two
independent weight-pairs, each positive on all five datasets.} The ten cells are not fully
independent, so the permutation figure supports a pattern already visible without it. Per-cell bootstrap intervals are wide and several include zero
(Appendix~\ref{app:matched}); the inference is the pattern across cells, not any single one.

\textbf{Two forms of the same effect.} On unbounded answer spaces, reasoning roughly halves the
number of distinct answers produced, a ratio of 0.43 to 0.65 across the three open-ended sets and
both scales, the diversity collapse stated in counts, with no derived statistic between the
observation and the claim. On bounded spaces, the key set is identical, four keys on GPQA-Diamond
and ten on MMLU-Pro for both arms, and collision rises anyway by \emph{more} than on mathematics
($+0.121$ to $+0.181$). What changes there is not the size of the answer space, but how mass is
distributed over it, and having both regimes is what makes the result robust in each direction.

\textbf{What the design excludes.} Three confounds proposed in recent work are excluded by
construction. \emph{Parameter scale}: \citet{tan2025consistent} report stable or growing self-consistent errors
with scale, so a comparison across size classes would inherit that trend, whereas
here the parameter count is identical within each pair and the effect appears at both 8B and 32B.
\emph{Family and training data}: within a pair they do not differ at all. \emph{Answer position}:
\citet{ding2026agree} shows a meaningful share of multiple-choice agreement is positional, but a
positional prior is a property of the weights and the prompt, both held fixed, and cannot explain
why one arm concentrates more mass than the other on the same four keys.

\textbf{On or off, not how much.} \citet{ghosal2025does} report that extending a reasoning model \emph{past} its natural stopping point raises the entropy of its answer distribution and lowers accuracy. That contrast varies the thinking budget rather than toggling thinking on or off, and measures entropy over all answers rather than collision among wrong ones. Their standard-thinking entropies are very low, 0.02 on MATH-500 and 0.23 on GSM-8K, so both results describe a concentrated distribution at the operating point we study; theirs adds that the concentration dissolves once the budget is forced beyond it.

\textbf{Generalization across eight models.} Comparing the four reasoning models against the
four non-reasoning models as unpaired groups (Table~\ref{tab:eightmodel}), on MATH-500, the separation is complete (16/16,
$p = 0.014$, the smallest attainable exact one-tailed Mann--Whitney probability at
$n_1 = n_2 = 4$); on GPQA-Diamond and MMLU-Pro the ordering is consistent but incomplete (14/16
each, $p = 0.057$), and we report those two as descriptive support rather than confirmation.
\textbf{The MATH-500 threshold was fixed in advance}: our pre-registered prediction stated
reasoning models would exceed 0.45 collision, and non-reasoning models fall below it, and all eight
fall on the predicted side, before and after cleaning. All four exceptions are Gemma-3 against Olmo-3-Think: between-family variation, which is exactly the noise the paired design removes.

\textbf{Errors are selected by accuracy: controlled.} The collision index is defined only on
problems where a model produces at least two errors, and the reasoning arm is substantially more
accurate, so the two arms contribute different problem sets; the selection is severe, with the
thinking arm contributing 25 problems on MATH-500 at 8B against the non-thinking arm's 117.
Repeating the comparison on the \textbf{matched subset} (problems on which \emph{each} arm
produces at least two errors) \textbf{nine of nine favor reasoning} (exact sign test
$p = 0.00195$), with median $\Delta = +0.124$ against $+0.133$ unmatched. AMC at 32B is dropped:
only seven problems qualify in both arms, below our pre-set minimum of ten. We treat the matched analysis as the confirmatory result; Table~\ref{tab:matched} gives the full comparison. Two residual threats:
AMC ($n = 83$) supplies both the smallest difference and the largest, so we report the median and
range rather than the mean; and Olmo-3-32B reaches 0.707 on MATH-500 against 0.480--0.509 for the
other reasoning models, so the unpaired comparison is carried in part by one model there, though
the paired result does not depend on it.

\section{What the concentration costs}
\label{sec:down}

\paragraph{The expected consequence, and why it does not appear.} If reasoning concentrates errors,
self-consistency should work less well on reasoning models. Measured as accuracy gained from one
sample to eight, that expectation appears emphatically confirmed in domain: reasoning models gain
$+0.011$ (from 0.948--0.964) against $+0.035$ for non-reasoning models (from 0.844--0.902), with no
overlap. \textbf{This comparison is invalid}, and we report it only because it is the comparison that a
reader will expect. Reasoning models have 0.036--0.052 of accuracy still available against
0.098--0.156 for non-reasoning models; expressing each gain as the fraction of available headroom
it captures, $(\mathrm{acc}_8 - \mathrm{acc}_1)/(1 - \mathrm{acc}_1)$, gives \textbf{0.250 against
0.249 in domain, a ratio of 1.00}. The threefold difference was the headroom ratio and nothing
else. Out of domain, where both arms are further from ceiling, reasoning converts nearly twice the
fraction non-reasoning does (0.163 against 0.091) and the raw gain ordering reverses. Accuracy gain
measures distance to ceiling, and the two arms differ in distance to ceiling by construction.

\begin{figure}[!tb]
\begin{center}
\includegraphics[width=4.393in]{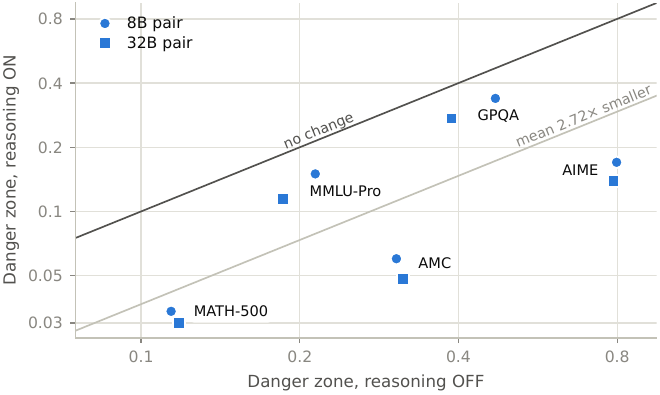}   
\end{center}
\caption{\textbf{Reasoning shrinks the set of problems voting can decide: 10 of 10 below the
diagonal, exact sign test $p = 0.00098$.} The danger zone with reasoning on against off, log--log,
one point per dataset--scale cell. The upper line is $y = x$, so a point below it is a cell where
reasoning shrank the zone; the lower line is the mean ratio, $2.72\times$. Per-cell values are in
Table~\ref{tab:danger}.}
\label{fig:danger}
\end{figure}

\paragraph{Two opposing effects, and why they cancel.} Voting can only be decided by answer
diversity on problems the model does not already get mostly right. Partition each problem by the
number of correct samples out of eight: where a majority are correct the plurality is correct
almost regardless of how errors are distributed, and where none are correct no distribution of
errors can help. Call the \textbf{danger zone} the fraction of problems with fewer than half the usable samples
correct: three or fewer of eight, zero included. Appendix~\ref{app:danger} states the rule formally. Within weights it falls in \textbf{ten of ten} dataset--scale cells (exact sign test $p = 0.00098$, and every cell's bootstrap interval excludes $1\times$), by between $1.39\times$ on GPQA-Diamond at 8B and $6.52\times$ on AMC at 32B,
with a mean of 0.136 reasoning-on against 0.369 reasoning-off: \textbf{$2.72\times$ smaller,
from identical weights} (Figure~\ref{fig:danger}; per-cell values in Table~\ref{tab:danger}). This solves
the puzzle. Reasoning concentrates errors, which makes voting worse \emph{within} the problems
where voting is decided by error diversity, and it shrinks that set by a factor of 2.7, because
the model is wrong less often. The second effect is larger, the two roughly cancel in aggregate
accuracy, and neither is visible if one only measures accuracy gain.

\paragraph{The rescue rate.} Inside the danger zone, does concentration cost what
Section~\ref{sec:mech} predicts? The \textbf{rescue rate} (the probability the plurality answer
is correct given only a minority of samples are) has no headroom term because it conditions on
the model already being mostly wrong. Pooled reasoning models rescue 0.098 of such problems
against 0.144; the ordering holds in 13 of 16 model pairings (exact $p = 0.100$), in both
within-weights pairs, and in three of the four cells with at least ten qualifying problems in both
arms ($p = 0.31$). \textbf{We report this as consistent with Section~\ref{sec:mech} and not as
confirmation of it}; the limitation is power, a direct consequence of the danger-zone result, since
reasoning models are too accurate to produce many minority-correct problems. One clean null runs
the other way, the probability that voting breaks a problem the model mostly gets right is
between 0.000 and 0.002 in every cell.

\paragraph{No confidence signal recovers what is left.} We test the obvious remedy across all six
signal families: one signal from each as a vote weight, plus a selector fitted on those same six,
against plain majority voting at $k = 8$, across eight models and five datasets:
\textbf{280 method--dataset--model combinations}, with Holm--Bonferroni applied once across the
whole family. \textbf{Not one combination shows a significant improvement.} At $\alpha = 0.05$, five of 280 reach
$p < 0.05$ uncorrected against 14.0 expected by chance; the smallest uncorrected $p$-value is
0.0161 and every Holm-adjusted value is 1.000. No choice of correction family changes the
conclusion, because the uncorrected count is already below chance expectation.
The mechanism of the null is that weighted voting almost never changes the answer: it agrees with
plain majority voting on 64{,}566 of 65{,}576 problem--method pairs (\textbf{98.5\% ties}), and
where the two differ the weighted vote is right 56.3\% of the time. The apparent effect also shrinks as power grows:
mean improvement runs $+0.0061$ at $n = 90$ against $+0.0015$ at $n = 500$.

\paragraph{What replicates.} \citet{bahuguna2026backfires} reports a 56.6\% backfire rate on
GPQA-Diamond for Qwen2.5-7B, and our non-reasoning Qwen3-8B has a danger zone of \textbf{47.0\%}
on the same benchmark. What we add
is the contrast his infrastructure could not reach: from the same weights with reasoning enabled it
falls to \textbf{33.9\%}. His paper states that the agreement gate's failure ``remains without a
mechanism''; Section~\ref{sec:mech} and this section supply one: agreement fails because reasoning
concentrates the errors it would need to discriminate, and it fails on fewer problems because
reasoning also makes fewer errors.

\paragraph{Threats.} The family covers vote weighting, not selection; a method that filters the
$k$ samples on confidence is outside it. Our budget fixed $k = 8$ where self-consistency is often
run higher, so while the direction of the danger-zone result is unlikely to reverse, its magnitude
is specific to this $k$. A second concentration measure, the rate of unanimously wrong problems,
was withdrawn as not headroom-free.

\section{What this does to confidence signals}
\label{sec:signals}

\paragraph{Agreement does not get worse when reasoning is switched off.} It gets slightly better,
consistently across both pairs and both splits: at 8B, 0.918 against 0.952 in domain and 0.805
against 0.869 out of domain; at 32B, 0.939 against 0.956 and 0.830 against 0.893. Four of four,
with the two out-of-domain effects agreeing to within 0.001 across a fourfold difference in
parameter count. Three caveats belong in the text rather than a footnote: per dataset the
non-thinking arm wins on only three of five, no individual dataset's confidence intervals separate
favorably, and macro-averaged \emph{between} models rather than within pairs the reasoning models
are very slightly ahead. This is a within-weights effect, and we do not claim that it survives the
between-model comparison.

\paragraph{The guessing ceiling explains the multiple-choice numbers.} A confidence signal on a
multiple-choice benchmark is bounded above by chance: a model of accuracy $a$ on a $1/c$-option
benchmark that knows the answer on a fraction $K = (a - c)/(1 - c)$ of problems and guesses
uniformly on the rest has $\text{ceiling} = K/a + \tfrac{1}{2}(1 - K)\,c/a$, because lucky guesses
are indistinguishable from knowledge by any signal computed from the model's own output
(Appendix~\ref{app:ceiling}). Across all sixteen model--benchmark cells, \textbf{every observed
agreement AUROC respects this bound}, with $c$ taken from the cleaned answer space; the smallest
margin is 0.094. The clearest case is the one that looks worst without the correction: Gemma-3-12B
reaches only 0.605 agreement AUROC on GPQA-Diamond, which reads as a broken signal, but its accuracy on the cleaned
answer space is 0.386 against a chance rate of 0.25, so $K = 0.181$, \textbf{53.0\% of its
correct answers are lucky guesses}, and the ceiling is 0.735. This also sharpens the previous
paragraph: the reasoning arm sits \emph{further below its own ceiling} in four of four comparisons,
and the GPQA rows are decisive because the non-thinking arm wins \emph{despite having the lower
ceiling} (0.927 against 0.876 at 32B, 0.902 against 0.838 at 8B).

\paragraph{Comparability.} A within-weights comparison is only valid if both arms answer at comparable rates. All ten dataset--scale cells agree to within a 2\% tolerance fixed before the check was run, and every cell answers at or above 0.9287; at that size the residuals cannot account for AIME collision gaps of +0.122 and +0.097, still less danger-zone ratios above 4.5×. Per-arm rates and the direction of the residual are in Appendix~\ref{app:rates}.

\section{Length, revisited}
\label{sec:length}

\paragraph{Why our length baseline is not spurious.} The output length is a strong confidence signal in
our data, reaching 0.965 AUROC on AMC, and an objection has to be disposed of first.
\citet{santilli2025revisiting} show that when a confidence measure and the correctness function are both length-dependent, the resulting AUROC is systematically distorted. Our correctness functions are exact match, so the channel they identify cannot operate; Appendix~\ref{app:length} gives the argument in full. \textbf{One exposure, disclosed}: a response the grader cannot map to an offered
option is scored incorrect, and unmappable responses are systematically longer, materially so in the Gemma family. The consequence is
bounded and its direction known: it inflates the multiple-choice length AUROC at the Gemma end and
\emph{shrinks} the out-of-domain gaps below, and cannot touch the open-ended result.

\paragraph{Length is not a reasoning phenomenon.} \citet{devic2026trace} report that trace length
becomes a useful confidence signal \emph{after} reasoning post-training, and conclude that
post-training ``fundamentally alters the relationship between trace length and correctness.'' On
MATH-500 our in-domain length AUROCs run 0.838 to 0.929 across all eight models, and the ordering
interleaves the arms completely: the two are indistinguishable (0.872 mean for reasoning against
0.878 for non-reasoning), and \textbf{the highest value belongs to Gemma-3-27B, which has had no
reasoning post-training of any kind}. We do not read this as a contradiction: both results are
correct on their own data, and one variable reconciles them. \textbf{Answer format.}
\citet{devic2026trace} evaluate on ten knowledge and multiple-choice benchmarks where response
length barely varies; ours is open-ended competition mathematics, where length tracks difficulty
across a wide range. Reasoning post-training is not \emph{necessary} for length to be a strong
confidence signal; a task with real length variance is. We pre-registered the opposite prediction that the reasoning arm would show stronger length coupling, and the scorecard records it as a failure.

\paragraph{Format, and hedging as length in disguise.} Length's AUROC ranges from 0.757 to 0.965 on
open-ended mathematics and from 0.521 to 0.798 on multiple choice. \textbf{The multiple-choice band
contains the entire range \citet{devic2026trace}} report, across ten benchmarks and six models,
an external replication of our multiple-choice numbers on a disjoint model set, and the explanation
for why our headline figure is higher than anything in their study. The same confound reaches
signals that do not look like length: counting hedging expressions (``maybe'', ``wait'',
``perhaps'') gives 0.807 AUROC for Qwen3-8B on MATH-500, and after residualizing on output length,
it falls to \textbf{0.474}, below chance. Normalizing the identical feature by length rather than
residualizing it destroys it directly, 0.815 raw against 0.611 as a density. This explains a result
reported independently: \citeauthor{devic2026trace}'s Appendix G finds that counting a few common
``forking'' tokens performs comparably to trace length, and their Section~5.1 shows length correlates
above Spearman 0.8 with the count of the fifty highest-entropy tokens. Their finding is that hedge
counts work; ours is \emph{why}. They are length wearing a semantic costume.

\paragraph{Length is the least portable signal we measure.} It degrades out of domain in \textbf{all eight} models by between 0.079 and 0.220 (per-signal gaps in Table~\ref{tab:oodgaps}). Agreement
degrades in seven, the probe in six, and sequence log-probability \emph{improves} out of domain in
six of eight; length has the largest degradation in six of the eight models and is tied within
0.003 in the other two. \textbf{This is a different claim from \citeauthor{devic2026trace}'s
difficulty result} and a reviewer will conflate the two. They report that trace length degrades on
high-difficulty subsets, a gradient in difficulty within a distribution. Ours is a gradient
across distributions: the same signal, fitted in domain, applied to a benchmark it was not fitted
on. A signal can be stable across difficulty and unstable across distributions, and length is.

\section{Signal validity depends on the model}
\label{sec:modeldep}

\paragraph{A signal can reverse sign within the same weights.} Fitting a linear selector over six
signals, one per family, the weight assigned to answer-token log-probability is negative in \textbf{all
four} reasoning models ($-0.176$, $-0.379$, $-0.101$, $-0.249$) and positive in three of four
non-reasoning models ($+0.129$, $+0.298$, $+0.211$; $-0.080$ for Qwen3-32B non-thinking) Table~\ref{tab:coefs}. Within
weights the two pairs behave differently and we report both. \textbf{At 8B the sign inverts}:
$-0.176$ with reasoning enabled against $+0.129$ with it disabled: with reasoning on, higher
confidence at the answer token predicts a \emph{wrong} answer. \textbf{At 32B the sign does not
invert, but the magnitude collapses 4.7-fold}, from $-0.379$ to $-0.080$, close to the selector
declining to use the signal at all. We claim the inversion at 8B only. This is the natural companion to Section~\ref{sec:mech}: if a chain commits the
model to an answer before the answer token is emitted, the log-probability there reports how firmly
the chain committed rather than how likely it is to be right, and among problems the model gets
wrong, firmer commitment is worse news. \citet{bahuguna2026backfires} reaches the same reading from
the other direction, describing the answer token as ``a near-deterministic readout of a chain that
has already committed.''

\paragraph{Answer-token entropy has no stable validity.} In domain the split is by family and complete; pooled across all five datasets it does not survive. Only Gemma-3 remains below chance (0.453, 0.458), while Olmo-3 moves from the bottom of the in-domain ordering to the top of the pooled one (0.598, 0.621). The useful statement is
therefore not that entropy is below chance for two families, which is true in domain and not out of
it, but that \textbf{the same signal ranges from 0.389 to 0.803 within a single model depending on
the benchmark}. Entropy computed over the $k$ \emph{answers} rather than over answer tokens is by contrast strong and stable everywhere, macro AUROC 0.828 to 0.879: two quantities both called ``entropy'' behave completely differently, and the one that works is computed over the sampled answer distribution.

\paragraph{Ranking mostly survives transfer; calibration does not.} Fitting a selector on one
family and evaluating on another, five of six off-diagonal AUROCs lie between 0.806 and 0.864 (Table~\ref{tab:transfer})
against diagonals of 0.819 to 0.870. The exception is Gemma-3 fitted and applied to Qwen3, at
0.728, and no transferred selector out-ranks the family it is evaluated on. Calibration degrades
further: a selector fitted on Gemma-3 and applied to Olmo-3 records ECE \textbf{0.343} against
\textbf{0.097} for Olmo's own, a \textbf{3.5-fold} degradation in exactly the property a
practitioner would act on. A threshold chosen on one family does not survive transfer even when
the ordering does.

\paragraph{Combining six signals buys nothing that generalizes.} Fitting a selector over all
six and comparing against raw agreement, the strongest single signal (Table~\ref{tab:selector}): pooled over all five
datasets (including the benchmark it was fitted on) it gains a mean of $+0.0118$ and improves
six of eight models. Restricted to the four benchmarks it was \emph{not} fitted on, it gains
\textbf{$-0.0003$}: nothing to four decimal places, losing on five of eight models, worst at
$-0.039$. Its apparent in-domain advantage is fit, not signal. \citet{ding2026agree} reports the
same shape from the deployment side: a confidence-routed cascade across three model tiers is
dominated by the trivial policy of always using the mid tier, 0.36 against 0.40. A signal good
enough to rank is not automatically good enough to route.

\section{Conclusions and recommendations}
\label{sec:recs}

Within fixed weights, reasoning concentrates a model's errors and shrinks the set of problems where
answer diversity decides anything; the two oppose and roughly cancel, so aggregate accuracy shows
neither. Nothing downstream repairs the first: across 280 combinations no confidence weighting beat
a plain majority vote, and a signal's direction proved a property of the model, not of the signal.
Each item below is something this study got wrong, nearly got wrong, or found that others had; all
are cheap, and none requires additional inference.

\textbf{Report the answer rate beside every accuracy}, or a difference in coverage masquerades as a
difference in capability, and \textbf{report the token budget and truncation rate per dataset},
since a pooled figure hides exactly the cells where it binds. \textbf{Use a length-invariant
correctness function, and say that you have} \citep{santilli2025revisiting}, and \textbf{report the
length-only baseline beside any new signal, or residualize on length}, since a signal that does not
beat length has not been shown to measure anything length does not. \textbf{Never compare a
confidence AUROC across answer formats} without the guessing ceiling: 0.605 against a bound of
0.735 looks like a broken signal rather than a constrained one.

\textbf{When comparing reasoning to non-reasoning behavior, hold the weights fixed if you can};
every failure of the unpaired test in our data is cross-family, not reasoning. \textbf{Report
collision, not only agreement}, when reasoning models are involved: the two come apart precisely on
the models where self-consistency is most used. \textbf{Evaluate on a decision, not only on AUROC},
and \textbf{report a signal's direction with the model it was measured on}. \textbf{Distrust a
metric with a headroom term}: our most instructive error was a threefold difference that vanished
under headroom normalization. \textbf{Register the predictions}: five of our sixteen did not hold, and one
of those failures is now the finding of Section~\ref{sec:length}. 


\subsection*{AI use statement}


AI assistance was used to partially design some methodology... It was used to help interpret some results. We did not use generative AI for brainstorming; all research ideas and the conceptual framework were developed by the authors. We did not use generative AI to develop the conceptual framework, to formulate the mathematical
claims, to derive the guessing ceiling, or to propose or refine the pre-registered predictions. We did not use generative AI to
generate synthetic datasets or for translation; qualitative and thematic analysis is not applicable to this work.

AI assistance was also used to create and revise the figures,
to edit code, to draft and revise some sections of the manuscript, and to edit for readability and structure.

We have reviewed all AI-assisted work. Every number in the manuscript was
re-derived from the analysis tables and the released records; the released code was executed and
its outputs checked against the paper's tables. We take responsibility for the final content of this work,
including all elements produced with the aid of generative AI.

\subsection*{Reproducibility statement}

All numbers in this paper derive from 74{,}944 recorded generations under the protocol in
Section~\ref{sec:setup}: eight open-weight models, five public benchmarks, $k = 8$ samples per problem,
fixed sampling parameters, one seed, and exact-match grading. The statistical protocol is stated in
full in Section~\ref{sec:setup}: cluster bootstrap over problems, Holm--Bonferroni applied once across
the family of 280 tests, paired sign tests for within-weights claims and unpaired exact tests for
across-model claims. Answer rates and truncation rates accompany every accuracy (appendix), and all
multiple-choice statistics use the cleaned answer space defined in Section~\ref{sec:setup}, with both
variants reported. The sixteen pre-registered predictions and their verdicts, including the five that did not hold, are in the appendix.

The supplementary material contains the full pipeline (generation, grading, the twelve signals, the
selectors and the analysis) with its test suite, the analysis tables that every table and figure in
this paper reads, the per-figure plotted series as CSV, and per-sample records for all 74{,}944
generations. Each record carries the graded answer, the grader layer that produced it, output
length, finish reason and all twelve signal values, which is sufficient to recompute the collision
index, the danger zone, the rescue rate and the 280-combination comparison \emph{without a GPU}; the
included export script does exactly this, and reproduces Table~\ref{tab:collision} from the released
fields alone. Prompts and completions are withheld: they embed benchmark items verbatim, and
\citet{rein2024gpqa} ask that their questions not be published in plain text to limit contamination.
Problem identifiers are stable hashes, so the records join against the original benchmarks without
redistributing any of their content. Inference-relevant dependency versions are pinned exactly. Code
is released under the MIT license, the derived records under CC BY 4.0.

\subsection*{Ethics statement}

This work uses public benchmarks and open-weight models, involves no human subjects and releases no
new data collected from people. We see no ethical concerns specific to this study beyond those
general to work on language-model reliability, to which our recommendations in Section~\ref{sec:recs} are
intended to contribute.

\bibliography{iclr2027_conference}
\bibliographystyle{iclr2027_conference}

\appendix

\section{Models and datasets}
\label{app:setup}

\begin{table}[h]
\caption{The eight models. The two Qwen3 pairs share parameters, tokenizer, pre-training corpus and
post-training recipe exactly within each pair; only whether a reasoning chain is generated differs.
Short names in the first column are used in the tables below.}
\label{tab:models}
\begin{center}
\small
\begin{tabular}{lllcc}
\toprule
short name & model & family & reasoning & params \\
\midrule
\texttt{qwen\_small}          & Qwen3-8B (thinking)      & Qwen3  & yes & 8B \\
\texttt{qwen\_small\_nothink} & Qwen3-8B (non-thinking)  & Qwen3  & \textbf{no (same weights)} & 8B \\
\texttt{qwen\_large}          & Qwen3-32B (thinking)     & Qwen3  & yes & 32B \\
\texttt{qwen\_large\_nothink} & Qwen3-32B (non-thinking) & Qwen3  & \textbf{no (same weights)} & 32B \\
\texttt{olmo\_small}          & Olmo-3-Think-7B          & Olmo-3 & yes & 7B \\
\texttt{olmo\_large}          & Olmo-3-Think-32B         & Olmo-3 & yes & 32B \\
\texttt{gemma\_small}         & Gemma-3-12B-IT           & Gemma-3 & no & 12B \\
\texttt{gemma\_large}         & Gemma-3-27B-IT           & Gemma-3 & no & 27B \\
\bottomrule
\end{tabular}
\end{center}
\end{table}

\begin{table}[h]
\caption{The five benchmarks. Domain and answer format vary independently: GPQA-Diamond and
MMLU-Pro are both out-of-domain \emph{and} multiple choice, which would confound the two were it not
for AIME and AMC, which are out-of-domain and open-ended.}
\label{tab:datasets}
\begin{center}
\small
\begin{tabular}{lccl}
\toprule
dataset & problems & answer format & split \\
\midrule
MATH-500     & 500 & open-ended numeric     & in-domain \\
AIME         &  90 & open-ended numeric     & out-of-domain \\
AMC          &  83 & open-ended numeric     & out-of-domain \\
GPQA-Diamond & 198 & 4-option multiple choice  & out-of-domain \\
MMLU-Pro     & 300 & 10-option multiple choice & out-of-domain \\
\bottomrule
\end{tabular}
\end{center}
\end{table}

Olmo-3 is included partly because it is fully open (weights, training data and intermediate
checkpoints), so contamination questions are answerable for that family in a way they are not for
the others.

\section{The collision index: robustness checks and a related measure}
\label{app:matched}

The collision index is defined only on problems where a model produces at least two incorrect
answers, and the reasoning arm is substantially more accurate, so the two arms of a pair contribute
different problem sets to the average in Table~\ref{tab:collision}. The reasoning arm's set is both
smaller and concentrated on items it finds harder. Table~\ref{tab:matched} repeats the comparison on
the problems where \emph{each} arm produces at least two errors.

\begin{table}[h]
\caption{Collision on the matched subset. \textbf{Nine of nine favor reasoning} (exact sign test
$p = 0.00195$), median $\Delta = +0.124$ against $+0.133$ unmatched. AMC at 32B is dropped: only
seven problems meet the criterion in both arms, below the pre-set minimum of ten.}
\label{tab:matched}
\begin{center}
\small
\begin{tabular}{llccccr}
\toprule
dataset & pair & \multicolumn{3}{c}{problems (on / off / common)} & collision on / off & $\Delta$ \\
\midrule
MATH-500     & 8B  & 25 & 117 & 21 & 0.457 / 0.374 & $+0.084$ \\
MATH-500     & 32B & 26 & 103 & 19 & 0.453 / 0.297 & $+0.156$ \\
AIME         & 8B  & 28 &  79 & 28 & 0.263 / 0.130 & $+0.134$ \\
AIME         & 32B & 28 &  77 & 28 & 0.193 / 0.099 & $+0.095$ \\
AMC          & 8B  & 13 &  39 & 12 & 0.274 / 0.139 & $+0.135$ \\
GPQA-Diamond & 8B  & 100 & 135 & 94 & 0.771 / 0.642 & $+0.129$ \\
GPQA-Diamond & 32B &  83 & 120 & 77 & 0.746 / 0.649 & $+0.097$ \\
MMLU-Pro     & 8B  &  65 & 100 & 60 & 0.744 / 0.620 & $+0.124$ \\
MMLU-Pro     & 32B &  48 &  75 & 47 & 0.766 / 0.678 & $+0.088$ \\
\bottomrule
\end{tabular}
\end{center}
\end{table}

Two differences from the unmatched table are worth stating. The matched deltas are slightly
\emph{smaller} on the multiple-choice sets ($+0.088$ to $+0.129$ against $+0.121$ to $+0.181$) and
slightly \emph{larger} on AMC at 8B ($+0.135$ against $+0.035$), and the matched analysis loses one
comparison to the minimum-subset rule. We report both tables and treat the matched one as the confirmatory result.

Table~\ref{tab:collci} gives bootstrap intervals for the unmatched differences of Table~\ref{tab:collision}.

\begin{table}[h]
\caption{Cluster-bootstrap intervals for the collision difference, 20{,}000 resamples, paired over
problems. Six of the ten unmatched intervals and four of the nine matched intervals exclude zero;
all nineteen point estimates are positive. \textbf{The index is low-powered per cell by
construction}: it is defined only where an arm produces at least two errors, and the matched
version requires that of both arms, which is rare precisely because the reasoning arm is accurate.
Every multiple-choice cell, where 47 or more problems qualify, excludes zero in both analyses
except MMLU-Pro at 32B matched. \textbf{The inference in Section~\ref{sec:mech} is the consistency
of sign across cells and the two independent weight-pairs, not per-cell significance}; that is what
a sign test tests. The contrast with Table~\ref{tab:danger}, where every cell is individually
significant, is the expected consequence of conditioning: the mechanism is measured on a small
selected subset, its downstream consequence on every problem.}
\label{tab:collci}
\begin{center}
\footnotesize
\begin{tabular}{llrcrcc}
\toprule
& & \multicolumn{2}{c}{unmatched (Table~\ref{tab:collision})} & \multicolumn{2}{c}{matched (Table~\ref{tab:matched})} & \\
\cmidrule(lr){3-4}\cmidrule(lr){5-6}
dataset & pair & $\Delta$ & 95\% CI & $\Delta$ & 95\% CI & $n$ matched \\
\midrule
MATH-500     & 8B  & $+0.129$ & $[-0.037,\ +0.294]$ & $+0.084$ & $[-0.075,\ +0.238]$ & 21 \\
MATH-500     & 32B & $+0.190$ & $[+0.034,\ +0.345]$ & $+0.156$ & $[-0.021,\ +0.311]$ & 19 \\
AIME         & 8B  & $+0.122$ & $[-0.003,\ +0.259]$ & $+0.134$ & $[-0.011,\ +0.285]$ & 28 \\
AIME         & 32B & $+0.097$ & $[-0.012,\ +0.220]$ & $+0.095$ & $[-0.023,\ +0.221]$ & 28 \\
AMC          & 8B  & $+0.035$ & $[-0.135,\ +0.218]$ & $+0.135$ & $[-0.026,\ +0.308]$ & 12 \\
AMC          & 32B & $+0.430$ & $[+0.104,\ +0.746]$ & \multicolumn{2}{c}{dropped, $n = 7$} & 7 \\
GPQA-Diamond & 8B  & $+0.121$ & $[+0.053,\ +0.189]$ & $+0.129$ & $[+0.059,\ +0.199]$ & 94 \\
GPQA-Diamond & 32B & $+0.135$ & $[+0.059,\ +0.212]$ & $+0.097$ & $[+0.020,\ +0.173]$ & 77 \\
MMLU-Pro     & 8B  & $+0.147$ & $[+0.057,\ +0.235]$ & $+0.124$ & $[+0.039,\ +0.207]$ & 60 \\
MMLU-Pro     & 32B & $+0.180$ & $[+0.075,\ +0.287]$ & $+0.088$ & $[-0.018,\ +0.193]$ & 47 \\
\bottomrule
\end{tabular}
\end{center}
\end{table}

\begin{figure}[h]
\begin{center}
\includegraphics[width=\linewidth]{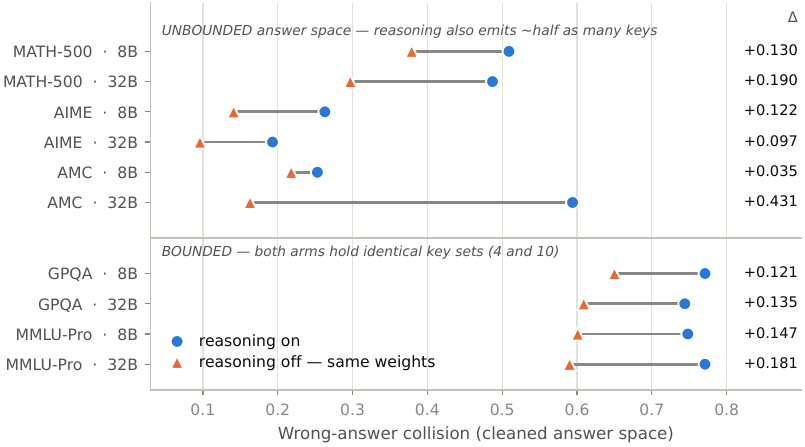}
\end{center}
\caption{\textbf{10 of 10 paired comparisons favor reasoning; exact sign test $p = 0.00098$.}
Collision among wrong answers with reasoning on against off, paired within fixed weights: each
line joins the two arms of one weight-pair on one benchmark, and every line points the same way.
The rule separates the two regimes of Section~\ref{sec:mech}: above it the answer space is
unbounded and reasoning also emits about half as many distinct keys; below it both arms hold
identical key sets of four and ten, so what changes is only how mass is distributed over them.
This is Table~\ref{tab:collision} drawn.}
\label{fig:collision}
\end{figure}

\paragraph{The unpaired eight-model comparison.} Table~\ref{tab:eightmodel} gives the collision
index for all eight models on the three benchmarks large enough to support the comparison
Section~\ref{sec:mech} reports. The paired design identifies the effect within fixed weights; this
one asks only whether the ordering holds across the wider model set, and it treats eight models as
eight independent observations when four of them are two weight-sharing pairs, so it is the weaker
of the two.

\begin{table}[h]
\caption{Collision among wrong answers for all eight models, on the three benchmarks large enough
to support the unpaired comparison. Values are the cleaned-answer-space index of
Section~\ref{sec:mech}. \emph{Dominances} counts the reasoning--non-reasoning model pairs, sixteen
per dataset, in which the reasoning model collides more; $p$ is the exact one-tailed Mann--Whitney
probability at $n_1 = n_2 = 4$. Bold marks the three cells where a Gemma-3 value exceeds an Olmo-3-Think value, breaking the
expected ordering; they yield four exceptions because one bold value exceeds both Olmo-3-Think
models. The 0.45 threshold Section~\ref{sec:mech} refers to
is prediction 7 in Table~\ref{tab:prereg}.}
\label{tab:eightmodel}
\begin{center}
\small
\begin{tabular}{lccc}
\toprule
model & MATH-500 & GPQA-Diamond & MMLU-Pro \\
\midrule
\multicolumn{4}{l}{\emph{reasoning}} \\
\texttt{qwen\_small}          & 0.509 & 0.771 & 0.748 \\
\texttt{qwen\_large}          & 0.487 & 0.744 & 0.771 \\
\texttt{olmo\_small}          & 0.480 & 0.662 & 0.624 \\
\texttt{olmo\_large}          & 0.707 & 0.718 & 0.730 \\
\midrule
\multicolumn{4}{l}{\emph{non-reasoning}} \\
\texttt{qwen\_small\_nothink} & 0.379 & 0.650 & 0.601 \\
\texttt{qwen\_large\_nothink} & 0.297 & 0.609 & 0.590 \\
\texttt{gemma\_small}         & 0.408 & 0.620 & \textbf{0.631} \\
\texttt{gemma\_large}         & 0.441 & \textbf{0.722} & \textbf{0.640} \\
\midrule
dominances & 16/16 & 14/16 & 14/16 \\
exact $p$  & 0.014 & 0.057 & 0.057 \\
\bottomrule
\end{tabular}
\end{center}
\end{table}

\paragraph{Relation to error semantic entropy.} \citet{aaai.Li} arrive independently at a measure of
the same distribution. They cluster a model's incorrect answers semantically and take the Shannon
entropy over the resulting clusters, $\mathrm{ESE} = -\sum_{k=1}^{K} p_k \log p_k$, where $p_k$ is
the share of errors falling in cluster $k$. Crossed with sampling accuracy, this separates four
failure states, and the one they call \emph{stably wrong}, where every sample is incorrect and
concentrated on a single wrong cluster, is the high-collision regime of Table~\ref{tab:collision}.
Entropy and the Herfindahl index of Section~\ref{sec:mech} are two standard concentration statistics
on one distribution, and they order it the same way; the convergence is worth recording because it
was reached from a different direction.

Their study differs from ours in what the measure is for and in what it is computed on. They use it
as an input to a confidence signal and align a model's verbal confidence to that signal by
reinforcement learning, reporting average expected calibration error reduced from 25.4\% to 6.4\%
and AUROC raised from 56.5\% to 67.3\% across six benchmarks; we hold the measure fixed as a
dependent variable and ask what the reasoning toggle does to it. Their samples are ten per question
from two instruction-tuned non-reasoning models, Llama-3-8B and Mistral-7B, on knowledge-QA
benchmarks, so the within-weights contrast is not available in their setting. And because their
answer spaces are open-ended text, they cluster answers semantically where our exact-match grading
partitions them without a modeling choice, a difference that bears on reproducibility rather than on
the direction of either result.

\section{The invalid comparison, in full}
\label{app:headroom}

Section~\ref{sec:down} reports that measuring the downstream consequence as a raw accuracy gain from
$k=1$ to $k=8$ produces a clean threefold difference that is an artifact. The per-model figures
are in Table~\ref{tab:headroom}.

\begin{table}[h]
\caption{Per-model accuracy at $k=1$ and $k=8$ on the in-domain benchmark, the headroom each model
has left, and the fraction of it that voting captures. The summary rows are the ranges
Section~\ref{sec:down} quotes. The final column is the point: the two arms differ almost
threefold in raw gain and not at all in the fraction of available headroom converted, and the
per-model values interleave rather than separating, so the equality is not an artifact of
averaging over groups.}
\label{tab:headroom}
\begin{center}
\small
\begin{tabular}{lccccc}
\toprule
model & $k=1$ & $k=8$ & gain & headroom & fraction converted \\
\midrule
\multicolumn{6}{l}{\emph{reasoning}} \\
\texttt{qwen\_small}          & 0.948 & 0.964 & $+0.016$ & 0.052 & 0.308 \\
\texttt{qwen\_large}          & 0.958 & 0.970 & $+0.012$ & 0.042 & 0.286 \\
\texttt{olmo\_small}          & 0.958 & 0.968 & $+0.010$ & 0.042 & 0.238 \\
\texttt{olmo\_large}          & 0.964 & 0.970 & $+0.006$ & 0.036 & 0.167 \\
\midrule
\multicolumn{6}{l}{\emph{non-reasoning}} \\
\texttt{qwen\_small\_nothink} & 0.844 & 0.894 & $+0.050$ & 0.156 & 0.321 \\
\texttt{qwen\_large\_nothink} & 0.854 & 0.898 & $+0.044$ & 0.146 & 0.301 \\
\texttt{gemma\_small}         & 0.852 & 0.880 & $+0.028$ & 0.148 & 0.189 \\
\texttt{gemma\_large}         & 0.902 & 0.920 & $+0.018$ & 0.098 & 0.184 \\
\midrule
reasoning, 4 models     & 0.948--0.964 & 0.964--0.970 & $\mathbf{+0.011}$ & 0.036--0.052 & \textbf{0.250} \\
non-reasoning, 4 models & 0.844--0.902 & 0.880--0.920 & $\mathbf{+0.035}$ & 0.098--0.156 & \textbf{0.249} \\
\bottomrule
\end{tabular}
\end{center}
\end{table}

\noindent Three times the benefit with reasoning switched off, with no overlap between the arms.
Reasoning models sit at 0.948--0.964 with between 0.036 and 0.052 of accuracy still available;
non-reasoning models sit at 0.844--0.902 with between 0.098 and 0.156 available. Expressing each
gain as the fraction of available headroom it captures:

\begin{center}
\small
\begin{tabular}{lccc}
\toprule
 & reasoning & non-reasoning & ratio \\
\midrule
in-domain             & 0.250 & 0.249 & \textbf{1.00} \\
pooled out-of-domain  & 0.163 & 0.091 & \textbf{0.56} \\
\bottomrule
\end{tabular}
\end{center}

\noindent Both arms convert almost exactly a quarter of what is available to them. Out of domain
the raw gain ordering also reverses ($+0.0391$ against $+0.0365$), and per dataset the raw
comparison splits three to two rather than eight to nothing: non-reasoning gains more on MATH-500,
AMC and MMLU-Pro, reasoning gains more on AIME ($+0.092$ against $+0.047$) and GPQA-Diamond
($+0.053$ against $+0.035$).

We report this at length because the same mistake is easy to make with any pair of models at
different accuracies, and because an earlier version of this paper made it.

\section{The danger zone, per cell}
\label{app:danger}

\paragraph{The rule, formally.} Let $k_p$ be the number of samples on problem $p$ that the grader
could map to an answer, and $c_p$ how many of those are correct. Problem $p$ is in the danger zone
when
\[
c_p \;<\; \lfloor k_p / 2 \rfloor,
\]
so at $k_p = 8$ the condition is $c_p \leq 3$, and unanimously wrong problems ($c_p = 0$) are
included. Problems with $k_p < 4$ are excluded, since a plurality over three or fewer samples
carries little information. The rescue rate conditions on the same set with the unanimously wrong
problems removed:
\[
\text{rescue rate} \;=\; \Pr\!\left[\,\text{plurality answer correct} \;\middle|\;
1 \leq c_p < \lfloor k_p / 2 \rfloor \,\right].
\]
The floor makes this marginally stricter than the word \emph{minority} when $k_p$ is odd: at
$k_p = 7$, three correct is excluded although three is short of a majority. That case arises only
where a sample was unmappable, and the rule applies identically to both arms.

\begin{table}[h]
\caption{The danger zone, the fraction of problems with fewer than half the usable samples correct.
Ten of ten favor reasoning (exact sign test $p = 0.00098$); mean 0.136 against 0.369,
\textbf{$2.72\times$ smaller}, from identical weights. Intervals are a paired cluster bootstrap over problems, 20{,}000 resamples. \textbf{Every lower
bound exceeds $1\times$}, which is the claim; upper bounds are unstable where the reasoning arm's
zone is small, since a single danger-zone problem out of 83 on AMC produces a ratio of 25.}
\label{tab:danger}
\begin{center}
\small
\begin{tabular}{llcccc}
\toprule
dataset & pair & reasoning on & reasoning off & ratio & 95\% CI \\
\midrule
MATH-500     & 8B  & 0.034 & 0.114 & $3.35\times$ & [2.21, 5.56] \\
MATH-500     & 32B & 0.030 & 0.118 & $3.93\times$ & [2.50, 7.00] \\
AIME         & 8B  & 0.170 & 0.798 & $4.69\times$ & [3.14, 8.50] \\
AIME         & 32B & 0.139 & 0.787 & $5.65\times$ & [3.71, 11.20] \\
AMC          & 8B  & 0.060 & 0.305 & $5.06\times$ & [2.62, 16.20] \\
AMC          & 32B & 0.048 & 0.314 & $6.52\times$ & [3.12, 25.00] \\
GPQA-Diamond & 8B  & 0.339 & 0.470 & $1.39\times$ & [1.17, 1.68] \\
GPQA-Diamond & 32B & 0.273 & 0.388 & $1.42\times$ & [1.20, 1.78] \\
MMLU-Pro     & 8B  & 0.150 & 0.214 & $1.43\times$ & [1.20, 1.77] \\
MMLU-Pro     & 32B & 0.114 & 0.186 & $1.64\times$ & [1.31, 2.16] \\
\bottomrule
\end{tabular}
\end{center}
\end{table}

\paragraph{Token cost.} Within fixed weights, reasoning spends $\mathbf{4.5\times}$ the tokens of
the same weights with thinking disabled: a problem-weighted mean of 6{,}324 against 1{,}405 at
$k = 1$, or $4.2\times$ at 8B and $4.9\times$ at 32B, with the per-model and per-dataset means in
Table~\ref{tab:tokens}. Self-consistency multiplies whatever that figure is by roughly eight. Comparing all four reasoning models against all four non-reasoning
models instead gives $6.6\times$ (7{,}418 against 1{,}128), but that number is composition, not
reasoning: Olmo-3-Think is the most verbose family in the study and Gemma-3 the tersest, and
neither belongs to a weight-pair. Within weights the $k=1 \rightarrow k=8$ gain is $+0.028$ for the
thinking arm against $+0.046$ for the non-thinking arm, from a lower base; between groups it is
$+0.027$ against $+0.036$. We draw no recommendation from any of these figures, because the two
arms are not interchangeable: the reasoning arm is between 0.10 and 0.15 more accurate at $k = 1$,
and a practitioner choosing between them is not choosing between equal-quality options at
different prices.

\begin{table}[h]
\caption{Mean output tokens per problem at $k=1$, by model and dataset, from which the ratios above
are computed. The last column is weighted by the number of problems in each dataset. The two
Qwen3 pairs are the identified comparison; the wide spread between Olmo-3-Think and Gemma-3, which
share no weights with anything, is why the eight-model ratio exceeds the within-weights one.}
\label{tab:tokens}
\begin{center}
\small
\begin{tabular}{lrrrrrr}
\toprule
model & MATH-500 & AIME & AMC & GPQA-D & MMLU-Pro & weighted mean \\
\midrule
\multicolumn{7}{l}{\emph{reasoning}} \\
\texttt{qwen\_small}          & 5{,}251 & 16{,}058 & 10{,}376 &  9{,}458 & 4{,}896 & 7{,}066 \\
\texttt{qwen\_large}          & 4{,}376 & 14{,}535 &  7{,}888 &  7{,}219 & 3{,}186 & 5{,}581 \\
\texttt{olmo\_small}          & 6{,}105 & 17{,}983 & 11{,}530 & 15{,}083 & 8{,}111 & 9{,}435 \\
\texttt{olmo\_large}          & 5{,}155 & 15{,}914 &  9{,}299 & 12{,}415 & 5{,}490 & 7{,}589 \\
\midrule
\multicolumn{7}{l}{\emph{non-reasoning}} \\
\texttt{qwen\_small\_nothink} & 1{,}200 &  5{,}005 &  2{,}269 &  1{,}922 & 1{,}136 & 1{,}674 \\
\texttt{qwen\_large\_nothink} &   878 &  3{,}124 &  1{,}479 &  1{,}363 &   723 & 1{,}135 \\
\texttt{gemma\_small}         &   859 &  1{,}527 &  1{,}221 &    957 &   563 &   877 \\
\texttt{gemma\_large}         &   771 &  1{,}911 &  1{,}137 &    845 &   488 &   824 \\
\bottomrule
\end{tabular}
\end{center}
\end{table}

\paragraph{The rescue rate, in full.} Pooled over datasets, reasoning models rescue 0.098 of
minority-correct problems against 0.144 for non-reasoning models. The ordering holds in 13 of 16
model pairings (exact $p = 0.100$) and in both within-weights pairs: 0.067 against 0.130 at 8B, and
0.149 against 0.201 at 32B. Restricting to the four cells with at least ten qualifying problems in
both arms, reasoning rescues less in three of four (exact $p = 0.31$). The limitation is power and
it is a direct consequence of the danger-zone result (Table~\ref{tab:danger}): reasoning models contribute 47--107
minority-correct problems against 121--154 for non-reasoning models. The measurement that would
settle this needs either more problems in the hard regime or a model pair closer in accuracy; we
have neither.

\begin{figure}[h]
\begin{center}
\includegraphics[width=4.435in]{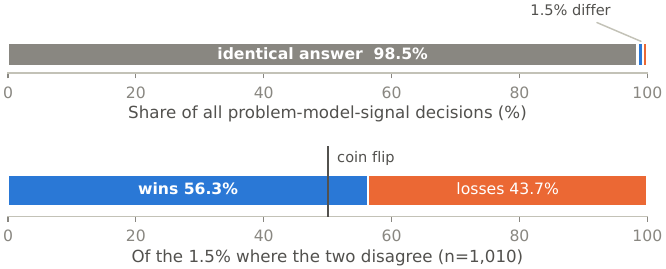}
\end{center}
\caption{What weighted voting actually does across the 280 combinations. It agrees with plain
majority voting on 64{,}566 of 65{,}576 problem--method pairs (98.5\%); where it differs it is
right 56.3\% of the time. The win rate in the differing cases fell monotonically as models were
added to the study: 58.9\%, then 56.6\%, then 56.3\%.}
\label{fig:decisions}
\end{figure}

\paragraph{The $p$-value distribution, in full.} Table~\ref{tab:pdist} gives all 280 raw
$p$-values by bin; the complete rows, raw and Holm-adjusted, are in the supplementary tables.

\begin{table}[h]
\caption{Distribution of the 280 raw $p$-values, against what a uniform null would produce. The
median raw value is 1.000, and every Holm-adjusted value is 1.000.}
\label{tab:pdist}
\begin{center}
\small
\begin{tabular}{lrr}
\toprule
raw $p$ & observed & expected under a uniform null \\
\midrule
$[0,\ 0.01)$      &   0 &   2.8 \\
$[0.01,\ 0.05)$   &   5 &  11.2 \\
$[0.05,\ 0.10)$   &   2 &  14.0 \\
$[0.10,\ 0.25)$   &  10 &  42.0 \\
$[0.25,\ 0.50)$   &  34 &  70.0 \\
$[0.50,\ 1.00]$   & 229 & 140.0 \\
\midrule
total             & 280 & 280 \\
\bottomrule
\end{tabular}
\end{center}
\end{table}

\noindent \textbf{The shift toward one is not additional evidence for the null.} It is
discreteness: weighted voting agrees with plain majority voting on 98.5\% of problem--method pairs,
so most combinations yield too few discordant decisions for the exact test to move, and a great
many return 1.000 exactly. The tests are therefore conservative, and the substantive evidence is
the decision-level breakdown (Figure~\ref{fig:decisions}) rather than the $p$-values: when the two
rules do differ, the weighted vote is right 56.3\% of the time. The five smallest raw values span
four models, three datasets and three methods, with nothing concentrating in any of them.

\section{Agreement within pairs, and the guessing ceiling}
\label{app:ceiling}

This appendix gives the per-pair agreement numbers behind Section~\ref{sec:signals} and the
derivation of the guessing ceiling that section applies. Table~\ref{tab:pairs} reports agreement
AUROC pooled within each weight-pair and split; $\Delta$ is the non-thinking arm minus the thinking arm, so a positive value favors reasoning off, as it does in all four rows.

\begin{table}[h]
\caption{Agreement AUROC pooled within each weight-pair and split. $\Delta$ is reasoning off minus
reasoning on.}
\label{tab:pairs}
\begin{center}
\small
\begin{tabular}{llccc}
\toprule
pair & split & reasoning on & reasoning off & $\Delta$ \\
\midrule
8B  & in-domain      & 0.918 & 0.952 & $+0.034$ \\
8B  & out-of-domain  & 0.805 & 0.869 & $+0.064$ \\
32B & in-domain      & 0.939 & 0.956 & $+0.017$ \\
32B & out-of-domain  & 0.830 & 0.893 & $+0.063$ \\
\bottomrule
\end{tabular}
\end{center}
\end{table}

\begin{figure}[h]
\begin{center}
\includegraphics[width=\linewidth]{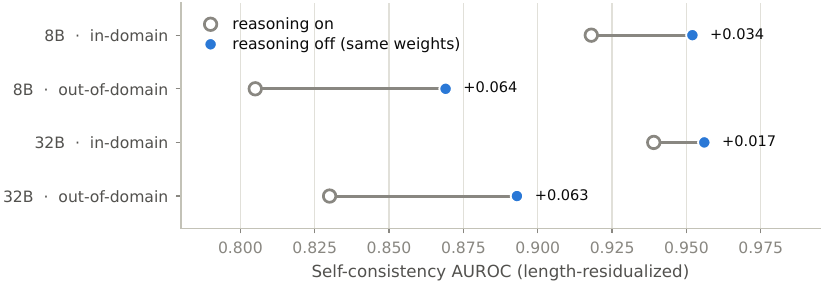}
\end{center}
\caption{Agreement AUROC with reasoning on against off, pooled within each weight-pair and split.
This is Table~\ref{tab:pairs} drawn.}
\label{fig:pairs}
\end{figure}

\paragraph{Derivation of the guessing ceiling.} Let a model have accuracy $a$ on a benchmark with
$1/c$ options, so $c$ is the chance rate. Assume it knows the answer on a fraction $K$ of problems
and guesses uniformly on the rest. Then $a = K + (1-K)c$, giving
\[
K = \frac{a - c}{1 - c}.
\]
Of the correct answers, a fraction $K/a$ come from knowledge and the remainder $(1-K)c/a$ from lucky
guesses. A signal computed from the model's own output can in principle rank every known-answer case
above every incorrect one, but it cannot distinguish a lucky guess from an unlucky one: those
contribute at chance. The bound is therefore
\[
\text{ceiling} \;=\; \frac{K}{a} \;+\; \tfrac{1}{2}\,\frac{(1-K)\,c}{a}.
\]
\textbf{One assumption, and its direction.} The bound assumes a knows-or-guesses model. Real partial
knowledge lowers $K$ further, which lowers the ceiling and pushes observed values \emph{closer} to
it. The error introduced by the assumption therefore runs in the direction that makes the bound
harder to satisfy, not easier.

Across all sixteen model--benchmark cells every observed agreement AUROC respects the bound, with
$c$ taken from the cleaned answer space: exactly four keys on GPQA-Diamond and exactly ten on
MMLU-Pro for all eight models. The smallest margin is 0.094. In both pairs and on both
multiple-choice datasets the reasoning arm sits further below its own ceiling than the non-thinking
arm; the GPQA rows are the decisive ones, because the non-thinking arm wins \emph{despite having the
lower ceiling}: 0.927 against 0.876 at 32B and 0.902 against 0.838 at 8B. A ceiling effect cannot
explain a result that runs against the ceiling.

\paragraph{The bound survives the uncertainty.} The ceiling is compared against point estimates
above; Table~\ref{tab:ceilcheck} adds the cluster-bootstrap interval for each agreement AUROC. In
none of the sixteen cells does the upper end of the interval reach the ceiling, so the result does
not depend on treating the AUROCs as exact. The tightest case is \texttt{olmo\_large} on
GPQA-Diamond, whose interval tops out at 0.850 against a ceiling of 0.895.

\begin{table}[h]
\caption{Agreement AUROC with its 95\% cluster-bootstrap interval against the guessing ceiling, for
all sixteen multiple-choice cells. Accuracy and chance rate are taken from the cleaned answer
space. Margin is ceiling minus the point estimate, computed from unrounded values, so it can differ by
0.001 from the difference of the two rounded columns beside it.}
\label{tab:ceilcheck}
\begin{center}
\small
\begin{tabular}{llccccc}
\toprule
model & dataset & acc & AUROC & 95\% CI & ceiling & margin \\
\midrule
\texttt{qwen\_small}          & GPQA-D & 0.629 & 0.742 & [0.686, 0.796] & 0.902 & 0.159 \\
\texttt{qwen\_large}          & GPQA-D & 0.695 & 0.773 & [0.714, 0.828] & 0.927 & 0.154 \\
\texttt{olmo\_small}          & GPQA-D & 0.550 & 0.745 & [0.691, 0.794] & 0.863 & 0.118 \\
\texttt{olmo\_large}          & GPQA-D & 0.613 & 0.800 & [0.757, 0.850] & 0.895 & \textbf{0.094} \\
\texttt{qwen\_small\_nothink} & GPQA-D & 0.508 & 0.735 & [0.682, 0.789] & 0.838 & 0.103 \\
\texttt{qwen\_large\_nothink} & GPQA-D & 0.574 & 0.772 & [0.723, 0.822] & 0.876 & 0.104 \\
\texttt{gemma\_small}         & GPQA-D & 0.386 & 0.605 & [0.543, 0.663] & 0.735 & 0.130 \\
\texttt{gemma\_large}         & GPQA-D & 0.481 & 0.668 & [0.605, 0.723] & 0.820 & 0.152 \\
\midrule
\texttt{qwen\_small}          & MMLU-Pro & 0.833 & 0.723 & [0.665, 0.794] & 0.989 & 0.266 \\
\texttt{qwen\_large}          & MMLU-Pro & 0.875 & 0.771 & [0.701, 0.848] & 0.992 & 0.221 \\
\texttt{olmo\_small}          & MMLU-Pro & 0.744 & 0.842 & [0.792, 0.892] & 0.981 & 0.139 \\
\texttt{olmo\_large}          & MMLU-Pro & 0.845 & 0.837 & [0.768, 0.898] & 0.990 & 0.153 \\
\texttt{qwen\_small\_nothink} & MMLU-Pro & 0.760 & 0.852 & [0.805, 0.893] & 0.982 & 0.131 \\
\texttt{qwen\_large\_nothink} & MMLU-Pro & 0.813 & 0.868 & [0.819, 0.917] & 0.987 & 0.119 \\
\texttt{gemma\_small}         & MMLU-Pro & 0.733 & 0.867 & [0.828, 0.904] & 0.980 & 0.112 \\
\texttt{gemma\_large}         & MMLU-Pro & 0.775 & 0.799 & [0.744, 0.858] & 0.984 & 0.185 \\
\bottomrule
\end{tabular}
\end{center}
\end{table}

\section{Answer rates and truncation}
\label{app:rates}

Comparing a thinking arm against a non-thinking arm is only meaningful if the two produce answers at
comparable rates. If one condition fails to emit a parseable answer more often than the other, a
difference in accuracy or in signal quality can be an artifact of the grader's coverage rather than
a property of the model. We adopt the rule from \citet{bahuguna2026backfires}, who enforced it
against his own pre-registered result. Below, \texttt{answer} is the fraction of samples yielding an
answer the grader could extract and, on multiple choice, map to an offered option; \texttt{trunc} is
the fraction whose generation stopped because it reached the 32{,}768-token budget.

\begin{table}[h]
\caption{Per-model answer, no-answer, unmappable, truncation and accuracy rates. Every cell answers
at or above 0.9287. The four bold cells are the Gemma-3 multiple-choice rows discussed in
Section~\ref{sec:length}: the model returns a computed value rather than an option letter, and those
responses are excluded from the answer space rather than counted as distinct answers. The
\texttt{acc} column is over all samples, with unmappable responses scored incorrect; accuracy on the
cleaned answer space is \texttt{acc} divided by \texttt{answer}, so \texttt{gemma\_small} on
GPQA-Diamond reads $0.3586/0.9287 = 0.386$, the figure the guessing-ceiling calculation uses.}
\label{tab:rates}
\begin{center}
\footnotesize
\begin{tabular}{llrccccc}
\toprule
model & dataset & $n$ & answer & no\_ans & unmap & trunc & acc \\
\midrule
\texttt{qwen\_small} & MATH-500     & 4000 & 0.9955 & 0.0045 & 0.0000 & 0.0055 & 0.9573 \\
\texttt{qwen\_small} & AIME         &  720 & 0.9222 & 0.0778 & 0.0000 & 0.0917 & 0.7222 \\
\texttt{qwen\_small} & AMC          &  664 & 0.9849 & 0.0151 & 0.0000 & 0.0226 & 0.8961 \\
\texttt{qwen\_small} & GPQA-Diamond & 1584 & 0.9987 & 0.0000 & 0.0013 & 0.0019 & 0.6282 \\
\texttt{qwen\_small} & MMLU-Pro     & 2400 & 1.0000 & 0.0000 & 0.0000 & 0.0021 & 0.8333 \\
\midrule
\texttt{qwen\_large} & MATH-500     & 4000 & 0.9968 & 0.0032 & 0.0000 & 0.0037 & 0.9593 \\
\texttt{qwen\_large} & AIME         &  720 & 0.9403 & 0.0597 & 0.0000 & 0.0792 & 0.7625 \\
\texttt{qwen\_large} & AMC          &  664 & 0.9880 & 0.0120 & 0.0000 & 0.0166 & 0.9322 \\
\texttt{qwen\_large} & GPQA-Diamond & 1584 & 0.9956 & 0.0000 & 0.0044 & 0.0019 & 0.6919 \\
\texttt{qwen\_large} & MMLU-Pro     & 2400 & 0.9996 & 0.0000 & 0.0004 & 0.0004 & 0.8742 \\
\midrule
\texttt{olmo\_small} & MATH-500     & 4000 & 0.9965 & 0.0035 & 0.0000 & 0.0055 & 0.9593 \\
\texttt{olmo\_small} & AIME         &  720 & 0.9097 & 0.0903 & 0.0000 & 0.1139 & 0.7083 \\
\texttt{olmo\_small} & AMC          &  664 & 0.9654 & 0.0346 & 0.0000 & 0.0422 & 0.9036 \\
\texttt{olmo\_small} & GPQA-Diamond & 1584 & 0.9937 & 0.0000 & 0.0063 & 0.0126 & 0.5461 \\
\texttt{olmo\_small} & MMLU-Pro     & 2400 & 1.0000 & 0.0000 & 0.0000 & 0.0088 & 0.7442 \\
\midrule
\texttt{olmo\_large} & MATH-500     & 4000 & 0.9972 & 0.0027 & 0.0000 & 0.0030 & 0.9617 \\
\texttt{olmo\_large} & AIME         &  720 & 0.9208 & 0.0792 & 0.0000 & 0.1028 & 0.7750 \\
\texttt{olmo\_large} & AMC          &  664 & 0.9789 & 0.0211 & 0.0000 & 0.0256 & 0.9428 \\
\texttt{olmo\_large} & GPQA-Diamond & 1584 & 0.9968 & 0.0000 & 0.0032 & 0.0057 & 0.6111 \\
\texttt{olmo\_large} & MMLU-Pro     & 2400 & 1.0000 & 0.0000 & 0.0000 & 0.0013 & 0.8454 \\
\midrule
\texttt{qwen\_small\_nothink} & MATH-500     & 4000 & 0.9922 & 0.0077 & 0.0000 & 0.0088 & 0.8445 \\
\texttt{qwen\_small\_nothink} & AIME         &  720 & 0.9403 & 0.0597 & 0.0000 & 0.0681 & 0.1944 \\
\texttt{qwen\_small\_nothink} & AMC          &  664 & 0.9669 & 0.0331 & 0.0000 & 0.0346 & 0.6431 \\
\texttt{qwen\_small\_nothink} & GPQA-Diamond & 1584 & 0.9874 & 0.0038 & 0.0088 & 0.0234 & 0.5013 \\
\texttt{qwen\_small\_nothink} & MMLU-Pro     & 2400 & 0.9946 & 0.0025 & 0.0029 & 0.0075 & 0.7558 \\
\midrule
\texttt{qwen\_large\_nothink} & MATH-500     & 4000 & 0.9978 & 0.0022 & 0.0000 & 0.0030 & 0.8608 \\
\texttt{qwen\_large\_nothink} & AIME         &  720 & 0.9597 & 0.0403 & 0.0000 & 0.0458 & 0.2250 \\
\texttt{qwen\_large\_nothink} & AMC          &  664 & 0.9834 & 0.0166 & 0.0000 & 0.0181 & 0.6355 \\
\texttt{qwen\_large\_nothink} & GPQA-Diamond & 1584 & 0.9905 & 0.0013 & 0.0082 & 0.0095 & 0.5682 \\
\texttt{qwen\_large\_nothink} & MMLU-Pro     & 2400 & 0.9975 & 0.0004 & 0.0021 & 0.0037 & 0.8108 \\
\midrule
\texttt{gemma\_small} & MATH-500     & 4000 & 0.9990 & 0.0010 & 0.0000 & 0.0010 & 0.8590 \\
\texttt{gemma\_small} & AIME         &  720 & 0.9944 & 0.0056 & 0.0000 & 0.0056 & 0.2028 \\
\texttt{gemma\_small} & AMC          &  664 & 0.9955 & 0.0045 & 0.0000 & 0.0045 & 0.5753 \\
\texttt{gemma\_small} & GPQA-Diamond & 1584 & \textbf{0.9287} & 0.0013 & \textbf{0.0701} & 0.0013 & 0.3586 \\
\texttt{gemma\_small} & MMLU-Pro     & 2400 & \textbf{0.9504} & 0.0000 & \textbf{0.0496} & 0.0000 & 0.6971 \\
\midrule
\texttt{gemma\_large} & MATH-500     & 4000 & 0.9992 & 0.0008 & 0.0000 & 0.0008 & 0.9012 \\
\texttt{gemma\_large} & AIME         &  720 & 0.9972 & 0.0028 & 0.0000 & 0.0028 & 0.2389 \\
\texttt{gemma\_large} & AMC          &  664 & 0.9955 & 0.0045 & 0.0000 & 0.0045 & 0.6852 \\
\texttt{gemma\_large} & GPQA-Diamond & 1584 & \textbf{0.9463} & 0.0000 & \textbf{0.0537} & 0.0006 & 0.4552 \\
\texttt{gemma\_large} & MMLU-Pro     & 2400 & 0.9812 & 0.0004 & 0.0183 & 0.0004 & 0.7604 \\
\bottomrule
\end{tabular}
\end{center}
\end{table}

\begin{table}[h]
\caption{Answer-rate parity within the paired comparison. \textbf{Ten of ten cells fall within a
2\% tolerance fixed before the check was run}; the largest divergence is $-0.0194$, on AIME at 32B.}
\label{tab:parity}
\begin{center}
\small
\begin{tabular}{llccrccc}
\toprule
dataset & pair & answer ON & answer OFF & $\Delta$ & trunc ON & trunc OFF & verdict \\
\midrule
MATH-500     & 8B  & 0.9955 & 0.9922 & $+0.0033$ & 0.0055 & 0.0088 & ok \\
MATH-500     & 32B & 0.9968 & 0.9978 & $-0.0010$ & 0.0037 & 0.0030 & ok \\
AIME         & 8B  & 0.9222 & 0.9403 & $-0.0181$ & 0.0917 & 0.0681 & ok \\
AIME         & 32B & 0.9403 & 0.9597 & $-0.0194$ & 0.0792 & 0.0458 & ok \\
AMC          & 8B  & 0.9849 & 0.9669 & $+0.0181$ & 0.0226 & 0.0346 & ok \\
AMC          & 32B & 0.9880 & 0.9834 & $+0.0045$ & 0.0166 & 0.0181 & ok \\
GPQA-Diamond & 8B  & 0.9987 & 0.9874 & $+0.0114$ & 0.0019 & 0.0234 & ok \\
GPQA-Diamond & 32B & 0.9956 & 0.9905 & $+0.0051$ & 0.0019 & 0.0095 & ok \\
MMLU-Pro     & 8B  & 1.0000 & 0.9946 & $+0.0054$ & 0.0021 & 0.0075 & ok \\
MMLU-Pro     & 32B & 0.9996 & 0.9975 & $+0.0021$ & 0.0004 & 0.0037 & ok \\
\bottomrule
\end{tabular}
\end{center}
\end{table}

\paragraph{The residual divergence and its direction.} The two largest gaps are both on AIME and
both negative (Table~\ref{tab:parity}): the thinking arm answers slightly \emph{less} often, because it truncates more
(0.0917 against 0.0681 at 8B; 0.0792 against 0.0458 at 32B). Longer reasoning traces reach the token
budget more often. This is the expected direction, and it is worth stating what it can and cannot
do. It cannot manufacture the collision result: the two largest effects there are the AIME
comparisons at $+0.122$ and $+0.097$, against a difference of under 2\% in the denominator each is
computed from. It cannot manufacture the danger-zone result either, where the AIME ratios are
$4.69\times$ and $5.65\times$. It does mean the thinking arm's AIME statistics are computed on
marginally fewer samples, and Table~\ref{tab:rates} reports the counts so a reader can see the base.

\paragraph{A qualification to the token-budget claim.} On MATH-500 truncation runs between 0.0008
and 0.0088 across all eight models, the sub-1\% regime the budget was chosen for. \textbf{On AIME
it runs between 0.0028 and 0.1139}, an order of magnitude higher, with the four reasoning models at
0.068--0.114 and the non-reasoning models at 0.003--0.068. The budget is ample for MATH-500 and
merely adequate for AIME, and the difference tracks reasoning: the models that generate longer
traces are the ones that hit the ceiling. Any statement about truncation at 32{,}768 tokens should
name the dataset it refers to rather than pooling.

\section{Length: the mutual-bias argument and the transfer gaps}
\label{app:length}

\paragraph{Why exact-match grading closes the channel.} \citet{santilli2025revisiting} show that
when a confidence measure and the correctness function are both length-dependent, the resulting
AUROC is systematically distorted: errors in the grader correlate with the signal, and the signal is
credited for detecting the grader's mistakes. They demonstrate this with length itself as a
baseline: token count scores competitively under lexical correctness functions such as ROUGE-L
and BERTScore, and ranks worst under a length-invariant one. Their measured length correlations are
large: Spearman $-0.9$ for negative sequence probability, $+0.7$ for perplexity and for semantic
entropy without normalization. Our correctness functions are exact match, a numeric answer
compared after normalization on MATH-500, AIME and AMC, an option letter on GPQA-Diamond and
MMLU-Pro, and neither grader has any length sensitivity for a length-dependent error to correlate
with, so the channel cannot operate. This is a property of the grading design rather than a
robustness check applied afterwards.

\paragraph{The one exposure, in full.} Exact-match immunity has a boundary: a response the grader
cannot map to an offered option is scored incorrect, and if unmappability correlates with length the
mutual bias returns through that door. It does. Unmappable responses are systematically longer, and
materially so in the Gemma family: point-biserial $r = 0.475$ for \texttt{gemma\_small} on
MMLU-Pro and $0.280$ for \texttt{gemma\_large} on GPQA-Diamond, against essentially zero for the
models with negligible unmappable rates. The consequence is bounded and its direction is known: it
inflates the apparent multiple-choice length AUROC at the Gemma end of the range, and it
\emph{shrinks} the out-of-domain gaps in Table~\ref{tab:oodgaps} by raising the out-of-domain term. It cannot touch the open-ended result, which is measured where no option mapping exists.

\paragraph{The per-model length numbers.} Table~\ref{tab:lengthauroc} lists the in-domain length
AUROC for all eight models. One distinction the main text compresses is worth making here: the two
Gemma-3 models have had no reasoning post-training of any kind, while the Qwen3 non-thinking arms
share weights with reasoning models and differ from them only at decode time. The highest value in
the study belongs to the first group, which is what makes this a claim about post-training rather
than about the toggle.

\begin{table}[h]
\caption{In-domain length AUROC on MATH-500, ordered by value. Bold marks the two models with no
reasoning post-training of any kind; the Qwen3 non-thinking arms are also marked \emph{no} but
share weights with reasoning models.}
\label{tab:lengthauroc}
\begin{center}
\small
\begin{tabular}{llc}
\toprule
model & reasoning & in-domain length AUROC \\
\midrule
\texttt{gemma\_large}         & \textbf{no}  & \textbf{0.929} \\
\texttt{olmo\_small}          & yes & 0.897 \\
\texttt{gemma\_small}         & \textbf{no}  & 0.893 \\
\texttt{qwen\_small}          & yes & 0.868 \\
\texttt{qwen\_large}          & yes & 0.865 \\
\texttt{olmo\_large}          & yes & 0.857 \\
\texttt{qwen\_large\_nothink} & no  & 0.854 \\
\texttt{qwen\_small\_nothink} & no  & 0.838 \\
\bottomrule
\end{tabular}
\end{center}
\end{table}

\paragraph{The transfer gaps in full.} Table~\ref{tab:oodgaps} gives the in-domain minus pooled
out-of-domain gap for the four representative signals of Section~\ref{sec:length}. A positive value
is degradation, so the length column being positive in every row is what makes length is the
least portable signal we measure.

\begin{table}[h]
\caption{In-domain minus pooled out-of-domain AUROC, by signal and model. A positive value is degradation.}
\label{tab:oodgaps}
\begin{center}
\small
\begin{tabular}{lrrrr}
\toprule
model & agreement & probe & length & logprob \\
\midrule
\texttt{qwen\_small}          & $+0.122$ & $+0.039$ & $+0.129$ & $-0.141$ \\
\texttt{qwen\_large}          & $+0.113$ & $+0.042$ & $+0.111$ & $-0.146$ \\
\texttt{olmo\_small}          & $+0.100$ & $-0.084$ & $+0.161$ & $+0.000$ \\
\texttt{olmo\_large}          & $-0.007$ & $-0.152$ & $+0.079$ & $-0.027$ \\
\texttt{qwen\_small\_nothink} & $+0.114$ & $+0.053$ & $+0.111$ & $+0.013$ \\
\texttt{qwen\_large\_nothink} & $+0.094$ & $+0.069$ & $+0.127$ & $-0.010$ \\
\texttt{gemma\_small}         & $+0.132$ & $+0.161$ & $+0.180$ & $-0.076$ \\
\texttt{gemma\_large}         & $+0.167$ & $+0.211$ & $+0.220$ & $-0.101$ \\
\bottomrule
\end{tabular}
\end{center}
\end{table}

\begin{figure}[h]
\begin{center}
\includegraphics[width=\linewidth]{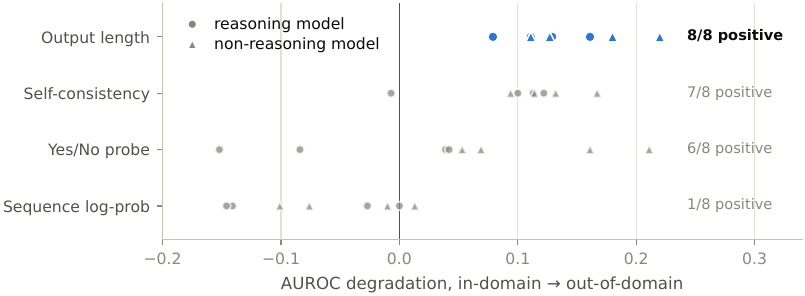}
\end{center}
\caption{In-domain minus out-of-domain AUROC, by signal and model. Length is the only signal that
degrades in all eight models. This is Table~\ref{tab:oodgaps} drawn.}
\label{fig:ood}
\end{figure}

\paragraph{On the residualization itself.} Our correction is a rank-space (Spearman)
residualization rather than the linear one \citet{vashurin2025line} propose. This is a difference of
family, not a reimplementation: they tested second- and third-degree polynomial fits and found no
improvement over linear, but did not test a monotone transform, and their own limitations anticipate
the gap: ``in tasks such as multi-step reasoning, where uncertainty may follow phase-specific
patterns, a linear fit may be insufficient.'' Reasoning traces are that case.

The correction is also signed, which is worth stating because it is a check on the instrument rather
than a result. Residualizing negative length on length drives it to 0.500 exactly, as it must. The
raw probe moves the other way, from 0.466 to 0.510, a signal that was being \emph{suppressed} by
its length correlation rather than inflated by it. \citet{santilli2025revisiting} measure the same
sign structure independently: $-0.9$ for negative sequence probability against $+0.7$ for
perplexity. On hedging, the count for \texttt{gemma\_small} falls from 0.467 to 0.148 after
residualization, alongside the \texttt{qwen\_small} fall from 0.807 to 0.474 reported in the main
text.

\section{Signal instability across models and datasets}
\label{app:modeldep}
In domain the split is by family and it is complete: on MATH-500 all four Qwen3 models score above
chance (0.586, 0.595, 0.617, 0.618) and all four Gemma-3 and Olmo-3 models below it (0.396, 0.402,
0.389, 0.478). Pooled across all five datasets, that separation does not survive (Table~\ref{tab:entropy}).

\begin{table}[h]
\caption{Mean answer-token entropy as a confidence signal, macro-averaged across all five datasets. Bold marks the two models that remain below chance.}
\label{tab:entropy}
\begin{center}
\small
\begin{tabular}{llc}
\toprule
model & family & macro AUROC \\
\midrule
\texttt{gemma\_small}         & Gemma-3 & \textbf{0.453} \\
\texttt{gemma\_large}         & Gemma-3 & \textbf{0.458} \\
\texttt{qwen\_large\_nothink} & Qwen3   & 0.517 \\
\texttt{qwen\_large}          & Qwen3   & 0.521 \\
\texttt{qwen\_small}          & Qwen3   & 0.542 \\
\texttt{qwen\_small\_nothink} & Qwen3   & 0.580 \\
\texttt{olmo\_large}          & Olmo-3  & 0.598 \\
\texttt{olmo\_small}          & Olmo-3  & 0.621 \\
\bottomrule
\end{tabular}
\end{center}
\end{table}

Only Gemma-3 remains below chance. Olmo-3 moves from the bottom of the in-domain ordering to the top
of the pooled one, because its entropy signal is weak in domain and strong on the two
competition-mathematics sets: \texttt{olmo\_large} runs 0.389 on MATH-500 and 0.803 on AIME. The
useful statement is therefore not ``entropy is below chance for two families,'' which is true in
domain and not out of it, but that the same signal ranges from 0.389 to 0.803 within a single model
depending on the benchmark, and from anti-correlated to useful across models on a single benchmark.

\begin{table}[h]
\caption{Cross-family transfer. Left: out-of-domain expected calibration error. Right: out-of-domain
AUROC. Rows are the family the selector was fitted on, columns the family it was evaluated on.
Ranking mostly transfers: five of six off-diagonal AUROCs lie between 0.806 and 0.864 against
diagonals of 0.819 to 0.870, the exception being Gemma-3 fitted and applied to Qwen3 at 0.728.
Calibration does not: Gemma-3 fitted, Olmo-3 evaluated gives ECE 0.343 against 0.097 for Olmo's
own, a 3.5-fold degradation.}
\label{tab:transfer}
\begin{center}
\small
\begin{tabular}{lccc@{\hskip 2.2em}ccc}
\toprule
& \multicolumn{3}{c}{ECE} & \multicolumn{3}{c}{AUROC} \\
\cmidrule(r){2-4}\cmidrule(l){5-7}
fitted on $\downarrow$ & Gemma-3 & Olmo-3 & Qwen3 & Gemma-3 & Olmo-3 & Qwen3 \\
\midrule
\textbf{Gemma-3} & 0.148 & \textbf{0.343} & 0.142 & 0.819 & 0.837 & 0.728 \\
\textbf{Olmo-3}  & 0.264 & 0.097 & 0.152 & 0.813 & 0.870 & 0.843 \\
\textbf{Qwen3}   & 0.189 & 0.097 & 0.099 & 0.806 & 0.864 & 0.843 \\
\bottomrule
\end{tabular}
\end{center}
\end{table}

The pattern in Table~\ref{tab:transfer} is not symmetric: Gemma-3 is both the worst family to fit on and the worst to be
evaluated on, with an ECE diagonal of 0.148 against 0.097 and 0.099 for the other two. This is the positive
reason to retain ECE despite its known defects as a headline metric. We do not use it to rank
signals; we use it to show that a signal can keep its ranking ability while losing any meaningful
relationship between its value and a probability.

\begin{table}[h]
\caption{Learned selector coefficients: the six signals the selector is fitted on, one from each
family, standardized so magnitudes are comparable within a row. Three patterns beyond the sign
reversal discussed in Section~\ref{sec:modeldep}. \emph{Self-consistency dominates every model},
which is why a fitted combination barely beats agreement alone (Table~\ref{tab:selector}); its
weight is also lower in every reasoning model (0.830--1.149) than in every non-reasoning model
(1.311--1.970), and roughly doubles within each weight-pair when reasoning is switched off. Both
\emph{answer entropy and hedging sit near zero throughout}: given trace length in the same model,
the selector declines to use the hedging count, which is the independent version of
Section~\ref{sec:length}'s finding that hedge counts are length in disguise. These are descriptive
observations from a single fit per model, not pre-registered tests: neither pattern appears in the
scorecard of Table~\ref{tab:prereg}, and both were found after the analysis was fixed. The fit
produces point estimates only, so no intervals are reported.}
\label{tab:coefs}
\begin{center}
\small
\begin{tabular}{lrrrrrr}
\toprule
model & probe & entropy & logprob & length & agreement & hedging \\
\midrule
\multicolumn{7}{l}{\emph{reasoning}} \\
\texttt{qwen\_small}          & $+0.776$ & $-0.028$ & $\mathbf{-0.176}$ & $-0.015$ & $+1.018$ & $-0.029$ \\
\texttt{qwen\_large}          & $+0.601$ & $+0.047$ & $\mathbf{-0.379}$ & $+0.218$ & $+1.149$ & $+0.160$ \\
\texttt{olmo\_small}          & $+0.492$ & $-0.047$ & $\mathbf{-0.101}$ & $+0.236$ & $+1.068$ & $+0.019$ \\
\texttt{olmo\_large}          & $+0.648$ & $+0.037$ & $\mathbf{-0.249}$ & $+0.172$ & $+0.830$ & $+0.037$ \\
\midrule
\multicolumn{7}{l}{\emph{non-reasoning}} \\
\texttt{qwen\_small\_nothink} & $+0.405$ & $-0.014$ & $\mathbf{+0.129}$ & $+0.113$ & $+1.943$ & $-0.029$ \\
\texttt{qwen\_large\_nothink} & $+0.319$ & $-0.025$ & $-0.080$          & $+0.292$ & $+1.970$ & $+0.029$ \\
\texttt{gemma\_small}         & $+0.737$ & $-0.072$ & $\mathbf{+0.298}$ & $+0.154$ & $+1.573$ & $+0.007$ \\
\texttt{gemma\_large}         & $+0.821$ & $-0.304$ & $\mathbf{+0.211}$ & $+0.542$ & $+1.311$ & $-0.008$ \\
\bottomrule
\end{tabular}
\end{center}
\end{table}

\begin{table}[h]
\caption{A fitted six-signal selector against raw agreement, the strongest single signal. The
distinction between the last two columns is the result: pooled over all five datasets (which
includes the benchmark the selector was fitted on) the selector gains $+0.0118$ and improves six
of eight models; restricted to the four benchmarks it was \emph{not} fitted on it gains $-0.0003$,
nothing to four decimal places, and loses on five of eight. Both $\Delta$ columns are computed from
unrounded AUROCs, so a $\Delta$ can differ by 0.001 from the difference of the two rounded columns
beside it; \texttt{qwen\_large} gains by under 0.0005 and is counted among the six.}
\label{tab:selector}
\begin{center}
\small
\begin{tabular}{lccrr}
\toprule
model & selector & agreement & $\Delta$ (all) & $\Delta$ (out-of-domain only) \\
\midrule
\texttt{qwen\_small}          & 0.897 & 0.869 & $+0.028$ & $+0.028$ \\
\texttt{qwen\_large}          & 0.888 & 0.887 & $+0.000$ & $-0.017$ \\
\texttt{olmo\_small}          & 0.899 & 0.884 & $+0.015$ & $+0.014$ \\
\texttt{olmo\_large}          & 0.912 & 0.890 & $+0.021$ & $+0.030$ \\
\texttt{qwen\_small\_nothink} & 0.892 & 0.878 & $+0.014$ & $-0.003$ \\
\texttt{qwen\_large\_nothink} & 0.898 & 0.898 & $-0.001$ & $-0.014$ \\
\texttt{gemma\_small}         & 0.875 & 0.855 & $+0.020$ & $-0.002$ \\
\texttt{gemma\_large}         & 0.842 & 0.845 & $-0.004$ & $-0.039$ \\
\midrule
\multicolumn{3}{r}{\textbf{mean}} & $\mathbf{+0.0118}$ & $\mathbf{-0.0003}$ \\
\multicolumn{3}{r}{models gaining} & 6 of 8 & 3 of 8 \\
\bottomrule
\end{tabular}
\end{center}
\end{table}

A six-signal learned combination (Table~\ref{tab:selector}), given in-distribution training data, does not out-perform
counting how often the samples agree. Its apparent in-domain advantage is fit, not signal, and
calibration is poor where it matters most: ECE 0.356 for \texttt{gemma\_large} on GPQA-Diamond.

\paragraph{The other entropy.} The entropy computed on the \emph{$k$ answers} rather than on the answer
tokens (the agreement-family signal) is strong and stable everywhere: macro AUROC 0.828 to
0.879 across all eight models, with no model below 0.599 on any dataset. Two quantities both called
``entropy'' behave completely differently, and the one that works is computed over the sampled
answer distribution rather than over the token distribution.

\section{Pre-registered prediction scorecard}
\label{app:prereg}

Recorded in writing before the corresponding numbers existed. Verdicts were re-checked against the
cleaned answer space; \textbf{all five collision-threshold verdicts are unchanged by the cleaning},
so the pre-registration is not sensitive to that analysis choice. Predictions 14 to 16 illustrate
the form: each fixes its threshold at Gemma-3-27B's already measured collision (0.431, 0.673 and
0.625 on the raw index) and predicts that Olmo-3-Think-7B, whose values did not yet exist, would
exceed it. Two of the three failed, one by 0.001.

\begin{table}[h]
\caption{Sixteen pre-registered predictions and their verdicts. Eleven passed; four failed,
and one was partial.}
\label{tab:prereg}
\begin{center}
\small
\begin{tabular}{rp{0.50\linewidth}l}
\toprule
\# & prediction & result \\
\midrule
 1 & Guessing ceiling bounds agreement AUROC & \textbf{PASS}, 16/16, no violations \\
 2 & Non-thinking 8B GPQA agreement $\leq 0.834$ & \textbf{PASS}, 0.741 \\
 3 & \dots and $\approx 0.07$ below the thinking arm's 0.743 & \textbf{FAIL}, moved 0.002 \\
 4 & MMLU-Pro ceilings near-equal, so any change is real signal & \textbf{PASS} as a test, within 0.006 \\
 5 & The reasoning arm shows stronger length coupling & \textbf{FAIL}, see Section~\ref{sec:length} \\
 6 & Length replicates on all large models, $>0.75$ on mathematics & \textbf{PASS}, 9/9 \\
 7 & Reasoning $>0.45$ MATH-500 collision, non-reasoning below & \textbf{PASS}, 8/8, and on cleaned data \\
 8 & The two Gemma models group together & \textbf{PASS} \\
 9 & Non-thinking 32B MMLU-Pro residualized agreement 0.82--0.84 & \textbf{PASS}, 0.820 \\
10 & The 32B pair wins on 4 of 5 datasets, losing AMC & \textbf{PARTIAL}, 3 of 5 \\
11 & Per-dataset confidence intervals will overlap & \textbf{PASS} \\
12 & Non-thinking 32B collision $<0.45$ / $<0.68$ / $<0.63$ & \textbf{PASS}, 0.297 / 0.609 / 0.590 \\
13 & AIME accuracy collapses toward 0.2 without reasoning & \textbf{PASS}, 0.762 $\rightarrow$ 0.225 \\
14 & Olmo-3-7B collision $>0.431$ on MATH-500 & \textbf{PASS}, 0.480 \\
15 & \dots $>0.673$ on GPQA-Diamond & \textbf{FAIL}, 0.662 \\
16 & \dots $>0.625$ on MMLU-Pro & \textbf{FAIL}, 0.624 \\
\bottomrule
\end{tabular}
\end{center}
\end{table}

Three of the five non-passes changed the paper. \textbf{Prediction 5} predicted that reasoning would
couple more strongly to length; it does not couple differently at all, and that null is now the
finding of Section~\ref{sec:length}: length is governed by answer format rather than by reasoning
post-training. \textbf{Predictions 15 and 16} each forced a claim about one model to be rewritten,
and their failure is why Section~\ref{sec:mech} reports the unpaired comparison as descriptive rather
than confirmatory. \textbf{Prediction 7} is the one worth reading closely in the other direction:
the 0.45 threshold was fixed in advance, and all eight models fall on the predicted side of it on
the largest dataset, in both the raw and cleaned analyses. That is stronger evidence than the
accompanying $p$-value, because the threshold was not chosen after seeing where the data fell.

\end{document}